\def\paperID{10}
\def\confName{ELVM}
\def\confYear{2025}

\def\paperTitle{Rethinking the Role of Spatial Mixing}

\def\authorBlock{
    George Cazenavette\textsuperscript{1}\thanks{Correspondence: \href{mailto:gcaz@mit.edu}{gcaz@mit.edu}} \qquad
    Joel Julin\textsuperscript{2} \qquad
    Simon Lucey\textsuperscript{3} \\\\\normalsize
    \textsuperscript{1}Massachusetts Institute of Technology \qquad \textsuperscript{2}Carnegie Mellon University \qquad \textsuperscript{3}University of Adelaide
}

\newif\ifreview 
\newif\ifarxiv \newcommand{\arxiv}{\arxivtrue}
\newif\ifcamera 
\newif\ifrebuttal 

\arxiv

\pdfoutput=1
\documentclass[10pt,twocolumn,letterpaper]{article}
\usepackage{multicol}

\ifreview \usepackage[review]{cvpr} \fi
\ifarxiv \usepackage[pagenumbers]{cvpr} \fi
\ifrebuttal \usepackage[rebuttal]{cvpr} \fi
\ifcamera \usepackage{cvpr} \fi

\usepackage{graphicx}	
\usepackage{amsmath}	
\usepackage{amssymb}	
\usepackage{booktabs}
\usepackage{times}
\usepackage{microtype}
\usepackage{epsfig}
\usepackage{caption}
\usepackage{float}
\usepackage{placeins}
\usepackage{color, colortbl}
\usepackage{stfloats}
\usepackage{enumitem}
\usepackage{tabularx}
\usepackage{xstring}
\usepackage{multirow}
\usepackage{xspace}
\usepackage{url}
\usepackage{subcaption}
\usepackage{xcolor}
\usepackage[hang,flushmargin]{footmisc}
\usepackage{graphicx}
\usepackage{amsmath}
\usepackage{amssymb}
\usepackage{booktabs}
\usepackage{duckuments}
\usepackage{algorithm}
\usepackage{algpseudocode}
\usepackage{amsmath}
\usepackage{amsfonts}
\usepackage{amssymb}
\usepackage{mathtools}
\usepackage{float}
\usepackage{makecell}

\usepackage[10pt]{moresize}

\usepackage{color,xcolor}
\usepackage{epsfig}
\usepackage{graphicx}
\usepackage{times}

\usepackage{comment}

\usepackage{pifont}

\usepackage{microtype}
\usepackage[font=smaller]{caption}
\usepackage{arydshln}
\usepackage{tabularx}
\usepackage{adjustbox}
\usepackage{array}
\usepackage{booktabs}
\usepackage{colortbl}
\usepackage{float,wrapfig}
\usepackage{makecell}
\usepackage{hhline}
\usepackage{multirow}
\usepackage{subcaption} %
\usepackage[percent]{overpic}
\usepackage{graphbox}
\usepackage{diagbox}
\usepackage{stmaryrd}

\usepackage{amsmath,amsfonts,amssymb}
\usepackage{bm}
\usepackage{nicefrac}
\usepackage{microtype}
\usepackage{dsfont}
\usepackage{changepage}
\usepackage{extramarks}
\usepackage{fancyhdr}
\usepackage{setspace}
\usepackage{soul}
\usepackage{xspace}
\usepackage{hhline}

\usepackage{url}

\usepackage{enumitem}
\usepackage[title]{appendix}

\ifcamera \usepackage[accsupp]{axessibility} \fi

\ifarxiv  \fi

\newcommand{\R}[1]{{%
    \textbf{%
        \ifstrequal{#1}{1}{\textcolor{red}{R#1}}{%
        \ifstrequal{#1}{2}{\textcolor{blue}{R#1}}{%
        \ifstrequal{#1}{3}{\textcolor{magenta}{R#1}}{%
        \ifstrequal{#1}{4}{\textcolor{teal}{R#1}}{%
                           \textcolor{cyan}{R#1}%
        }}}}%
    }%
}}

\definecolor{MyDarkBlue}{rgb}{0,0.08,1}
\definecolor{MyDarkGreen}{rgb}{0.02,0.6,0.02}
\definecolor{MyDarkRed}{rgb}{0.8,0.02,0.02}
\definecolor{MyDarkOrange}{rgb}{0.40,0.2,0.02}
\definecolor{MyPurple}{RGB}{111,0,255}
\definecolor{MyRed}{rgb}{1.0,0.0,0.0}
\definecolor{MyGold}{rgb}{0.75,0.6,0.12}
\definecolor{MyDarkgray}{rgb}{0.66, 0.66, 0.66}
\definecolor{MyDarkCyan}{rgb}{0.05, 0.55, 0.45}
\definecolor{MyBlack}{rgb}{0., 0., 0.}
\definecolor{MyMagenta}{rgb}{1., 0., 1.}
\definecolor{BerkeleyYellow}{RGB}{255,204,41}
\definecolor{BerkeleyLightBlue}{RGB}{94,146,221}
\definecolor{BkDarkBlue}{rgb}{.05,.07,.353}

\definecolor{MyDarkGray2}{rgb}{0.6, 0.6, 0.6}
\definecolor{MyRed}{rgb}{1.0,0.0,0.0}

\usepackage{xr-hyper}

\makeatletter
\newcommand*{\addFileDependency}[1]{
  \typeout{(#1)}
  \@addtofilelist{#1}
  \IfFileExists{#1}{}{\typeout{No file #1.}}
}

\makeatother
\newcommand*{\myexternaldocument}[1]{
    \externaldocument{#1}
    \addFileDependency{#1.tex}
    \addFileDependency{#1.aux}
}

\definecolor{cvprblue}{rgb}{0.21,0.49,0.74}
\usepackage[pagebackref,breaklinks,colorlinks,allcolors=cvprblue]{hyperref}
\usepackage[capitalize]{cleveref}
\crefname{section}{Sec.}{Secs.}
\crefname{table}{Table}{Tables}
\crefname{figure}{Fig.}{Figs.}

\ifarxiv \crefname{appendix}{App.}{Apps.}
\else \crefname{appendix}{Suppl.}{Suppls.} \fi

\unless\ifarxiv \myexternaldocument{_supplementary} \fi

\begin{document}
\title{\paperTitle}
\author{\authorBlock}
\maketitle

\begin{abstract}
Until quite recently, the backbone of nearly every state-of-the-art computer vision model has been the 2D convolution. At its core, a 2D convolution simultaneously mixes information across both the spatial and channel dimensions of a representation. Many recent computer vision architectures consist of sequences of isotropic blocks that disentangle the spatial and channel-mixing components. This separation of the operations allows us to more closely juxtapose the effects of spatial and channel mixing in deep learning. In this paper, we take an initial step towards garnering a deeper understanding of the roles of these mixing operations. Through our experiments and analysis, we discover that on both classical (ResNet) and cutting-edge (ConvMixer) models, we can reach nearly the same level of classification performance by \textit{only learning channel mixing} and leaving the spatial mixers at their random initializations. Furthermore, we show that models with random, fixed spatial mixing are naturally more robust to adversarial perturbations. Lastly, we show that this phenomenon extends past the classification regime, as such models can also decode pixel-shuffled images.
\end{abstract}
\vspace{-0.5cm}
\section{Introduction}
For the better part of the last two decades, cascades of learned convolutions have formed the backbone of nearly every innovation in the field of computer vision and pattern recognition. From distinguishing ten digits with LeNet \cite{lenet} to one thousand classes with AlexNet \cite{alexnet}, from going very deep with VGG \cite{vgg} to even deeper with InceptionNet \cite{inception}, the learned 2D convolution has served as the workhorse that ushered in the new era of visual learning. Convolutional neural networks (CNNs) learned to exploit the correlations across channels and between spatially close pixels to solve a plethora of tasks. 

\looseness=-1
Recently, isotropic networks (those in which the size of the representation stays fixed throughout) such as Vision Transformer \cite{visiontransformer}, Image GPT \cite{imagegpt}, MLP-Mixer \cite{mlpmixer}, and ConvMixer \cite{convmixer} have been grabbing the field's attention. 
These isotropic models consist of repeated blocks wherein each block consists of a spatial mixing operation (self-attention \cite{visiontransformer,imagegpt}, spatial MLP \cite{mlpmixer}, or depthwise convolution \cite{convmixer}) followed by one or more \textit{strictly} channel-mixing layers (1$\times$1 or ``pointwise'' convolutions). 
A strictly spatial mixing operation is defined here as any operation that is applied independently across each channel. Similarly, a strictly channel-mixing layer is applied independently across spatial pixel coordinates. 
The fact that all these recent state-of-the-art isotropic architectures spend a significant portion of their computation budget solely on channel-mixing parameters hints toward the possibility that channel mixing may have significantly more relative importance than previously appreciated.

\looseness=-1
It has been understood for some time that impressive performance can be obtained from networks whose weights are not all completely learned ~\cite{saxe2011random}. 
Before the success of AlexNet~\cite{alexnet} in training deep end-to-end networks, the use of random weights in early convolutional layers played an important role in training deeper CNNs. 
The approach was attractive as it promised faster training times, better generalisation, and the ability to learn deeper networks. 
Saxe et al.~\cite{saxe2011random} even offered strong theoretical motivations for why random filter weights in CNNs would offer good performance in terms of: (i) frequency selection, and (ii) translation invariance. 
One of the most notable works with respect to using random weights within networks can be found in the Extreme Learning Machines (ELMs) of Huang et al~\cite{huang2006extreme}. 
In its simplest form an ELM takes a proposed network architecture and randomly initializes all hidden weights, leaving only the final layer to learn. 
ELMs are advantageous as they allow for extremely deep networks and rapid train times (as only the final layer needs to be learned). 
It is widely understood, however, that these training strategies have a significant performance gap in relation to their current end-to-end learned counterparts. 

\looseness=-1
In this work, we leverage the separable convolution to compare the relative effects of spatial and channel mixing on the performance of deep neural networks. By leaving either the spatial-mixing or channel-mixing parameters frozen at their random initialization and only training the others, we can isolate the contributions of both types of parameters. Doing so reveals that learning the \textit{channel-mixing} parameters is \textit{far} more critical to the effectiveness of a deep model, and such models that only learn channel-mixing parameters perform nearly just as well as their fully-learned counterparts. 

We also show that models that only learn channel-mixing parameters are naturally more resistant to adversarial attacks than fully-learned models. This robustness can be further enhanced by then directly smoothing the random, frozen spatial-mixing filters.

Lastly, we show that this phenomenon extends beyond the regime of classification problems and find that models that only learn spatial mixing are even capable of learning to invert a permutation of pixels nearly just as well as fully-learned networks.

We hope that these gained insights as to the role and effectiveness of spatial and channel mixing in deep models will be of great value to the vision community and that they will shed light on which parts of deep models are most critical to learn while helping to inspire the intelligent design of future architectures.

\section{Related Work}
\paragraph{Isotropic Architectures} 
\looseness=-1
First popularized by the transformer \cite{vaswani2017attention}, isotropic architectures (those in which the size of the representation remains fixed throughout) have more recently made their way into the computer vision world. Image-GPT \cite{imagegpt} (based on the GPT language model \cite{imagegpt}) was able to model sequences of pixels and generate new images in raster fashion. Designed for discriminative tasks, the Vision Transformer \cite{visiontransformer} used transformer blocks to achieve state-of-the-art image classification results, but only after extensive pre-training. The MLP-Mixer \cite{mlpmixer} then built off Vision Transformer's success, but used a simple spatial-mixing MLP instead of the expensive self-attention module. While the MLP-Mixer still required extensive pre-training, the newest addition to the line of isotropic architectures, the ConvMixer \cite{convmixer}, replaces the self-attention or spatial-mixing MLP with a simple depthwise filter bank and achieves comparable results \textit{without} any pre-training. The main thing all these models have in common is their \textit{isotropic} structure. Here, ``isotropic'' refers to repeated blocks of operations in which the latent representation does not change in shape. Furthermore, in all of these models, the second half of each isotropic block is made up of one or more \textit{strictly-channel-mixing} layers. Such transformer-inspired isotropic models have recently come to be known as ``meta-formers'' \cite{yu2022metaformer}.
\paragraph{Separable Convolutions}
\looseness=-1
Instead of mixing across channel and spatial dimensions simultaneously, the separable convolution factorizes the operation into a depthwise and pointwise convolution. The depthwise convolution applies a disjoint set of depth-1 filters to each channel of the input independently while the pointwise convolution is simply a 1$\times$1 convolution (or linear projection) on the pixels of this intermediate representation. Separable convolutions are used extensively in many state-of-the-art architectures \cite{howard2017mobilenets,Zhang2018shuffle,Chollet2017sep} and have built-in implementations in deep learning libraries \cite{tensorflow2015-whitepaper}. Furthermore, a recent method \cite{dbouk2021generalized} converts a \textit{pre-trained} network with standard convolutions and converts them to separable ones, increasing the throughput and robustness of the model.
\paragraph{Lottery Ticket Hypothesis}
Recent work has uncovered an interesting phenomenon in deep neural networks wherein randomly initialized networks tend to contain much smaller sub-networks that can reach similar performance as the full network when both are trained to convergence. This has been dubbed the ``Lottery Ticket Hypothesis'' \cite{frankle2018lottery} with the analogy being that the optimal sub-network of randomly initialized weights has the winning ``ticket.'' In this work, we provide evidence of a similar phenomenon in very wide networks with fixed, randomly initialized spatial mixing layers wherein they reach the same performance as their fully-learned counterparts (Section \ref{sec:width}).
\paragraph{Fixed Spatial Mixing}
Some recent works have explored using fixed ``shift'' operations \cite{wu2018shift,chen2019all} or simple average pooling \cite{han2021learning,yu2022metaformer} as spatial-mixing operations. Our work significantly differs from these in several key areas. Rather than proposing a new architecture, our work seeks to exploit existing useful mixing pattern present at the network's random initialization (as inspired by the lottery ticket hypothesis \cite{frankle2018lottery}). Furthermore, these recently proposed models \cite{wu2018shift,chen2019all,han2021learning,yu2022metaformer} still \textit{learn} varying amounts of spatial-mixing layers whereas we analyze networks that do not learn \textbf{any} spatial-mixing parameters whatsoever. Our analysis seeks to illuminate which parameters must be learned and which can be left at their random initializations in architectures already in use today.

\begin{figure}
    \centering
    \includegraphics[width=\linewidth]{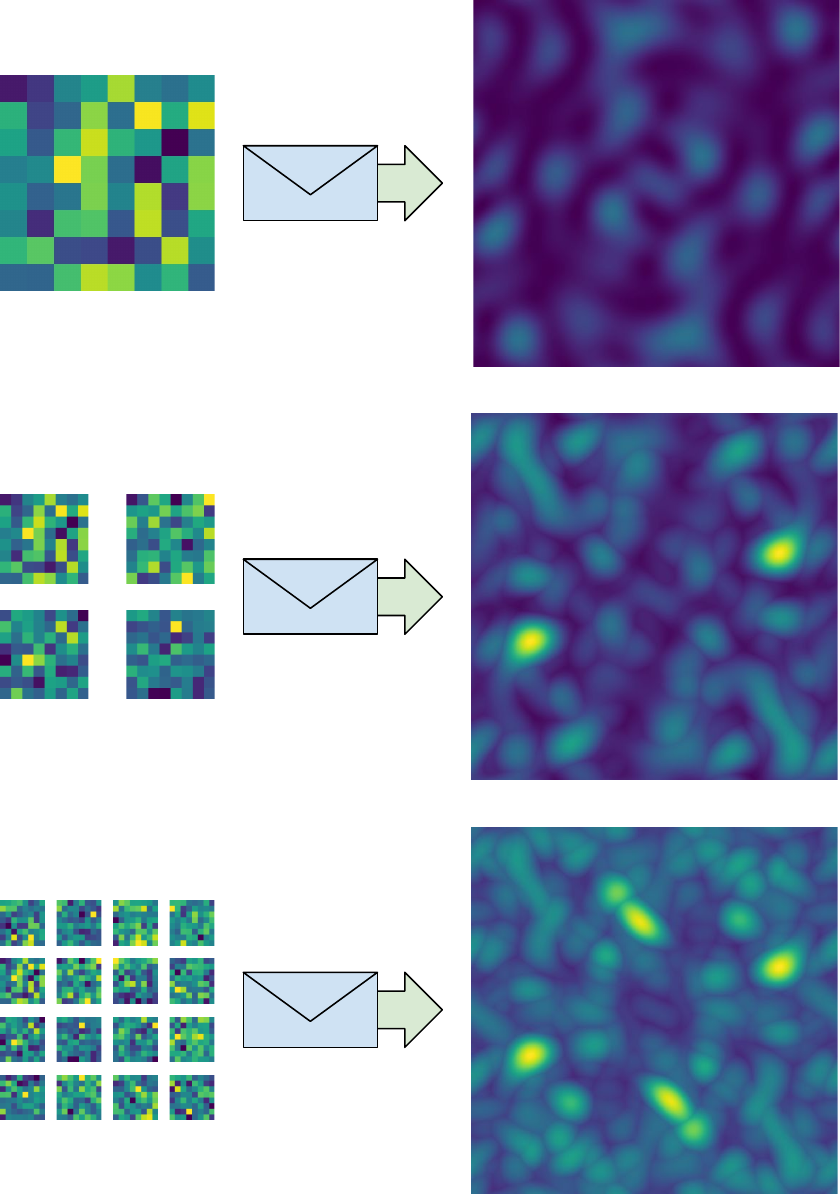}
    \caption{The spectral envelope of a bank of random filters rapidly grows as we add more uncorrelated filters to the set, allowing for quality feature extraction from natural images.}
    \vspace{-0.5cm}
    \label{fig:random-env}
\end{figure}
\section{Background and Setup}

Before our experiments, we provide some relevant exposition regarding the operations and architectures we use to illustrate the relative importance of spatial and channel mixing in deep neural networks.

\subsection{Spectral Coverage}\label{sec:spec}
Part of the efficacy of a spatial convolution can be attributed to the size of the \textit{spectral envelope} of its filter bank, or how much \textit{spectral coverage} it has. By spectral coverage, we mean the number of signal frequencies covered by the filter bank. By using multiple filters in tandem, the region of the envelope grows to be the union of each filter's individual coverage. 

Previous work \cite{saxe2011random, huang2006extreme} has shown that random filters make adequate visual feature extractors. Furthermore, motivated by decades of work in signal processing, it has been known that random filters have good spectral coverage. We hypothesize that these phenomena are related, and that the success of random filters is due in part to the size of their induced spectral envelope. As seen in Figure \ref{fig:random-env}, increasing the number of filters rapidly increases the bank's coverage in frequency space.\footnote{We visualize the spectral envelope by taking the max over the magnitude of the 2D fast Fourier transform (FFT) of each filter padded to be 512$\times$512. In our visualization the zero frequency component has been shifted to the center of the image.}

\looseness=-1
We hypothesize that random filters forming such a dense spectral envelope contributes to their ability to produce well-performing models. Intuitively, one could consider the random filters as extracting features from all different frequencies of input while the \textit{learned} channel-mixing parameters select linear combinations of these features that are actually useful.
\subsection{The 2D Convolution}\label{sec:conv}
\looseness=-1
As we begin our investigation into the importance of channel mixing, we focus on the simple yet revolutionary ResNet \cite{resnet} architecture. The original ResNet architecture is composed of blocks of 2D convolution operations. Given the number of in and out channels ($c_{in}, c_{out}$) and the size of the square kernel ($k$), we can represent the number of trainable parameters in the standard 2D convolution ($p_{conv}$) as 
\begin{equation}
    p_{conv} = c_{in}\cdot c_{out} \cdot k^2
\end{equation}
With the vanilla Conv2D, it is hard to analyze the respective importance of spatial and channel mixing since the two are entangled in a single linear operation. 

To disentangle the spatial and channel mixing \cite{howard2017mobilenets,Zhang2018shuffle,Chollet2017sep}, we separate the standard 2D convolution operation into a paired spatial-mixing (depthwise) and channel-mixing (1$\times$1 or pointwise) convolution. In a separable convolution, each filter of the depthwise convolution operates on just a single channel of the input. The pointwise convolution is simply a $1\times 1$ convolution, or a linear projection of each pixel.

After separating the standard 2D convolution into a depthwise and pointwise convolution, we can then introduce a \textit{depth multiplier} to increase the number of filters per input channel. Naturally, this also increases the number of channels in the pointwise filters since the intermediate representation will now have more channels. Given the number of in and out channels ($c_{in}, c_{out}$), the size of the square depthwise kernel ($k$), and the depth multiplier $d$, we can represent the number of trainable parameters in the separable 2D convolution ($p_{sep}$) as
\begin{equation}
    \begin{split}
        p_{sep} &= p_{depth} + p_{point}\\
        &= c_{in} \cdot k^2 \cdot d + c_{in} \cdot d \cdot c_{out}
    \end{split}
\end{equation}
By exploiting the inherent dependencies between nearby pixels, the standard 2D convolution---and its separable cousin---offer vision systems a cheap, yet effective way of extracting information from images.

\begin{figure*}[t]
    \centering
    \small
    \begin{tabular}{cccccc}
    &&&\large{Separable ResNet}\\
    & CIFAR-10 && CIFAR-100 && ImageNet\\
        \rotatebox[origin=c]{90}{Validation Acc. \%} &\hspace{-.5cm} \includegraphics[align=c,width=0.29\linewidth]{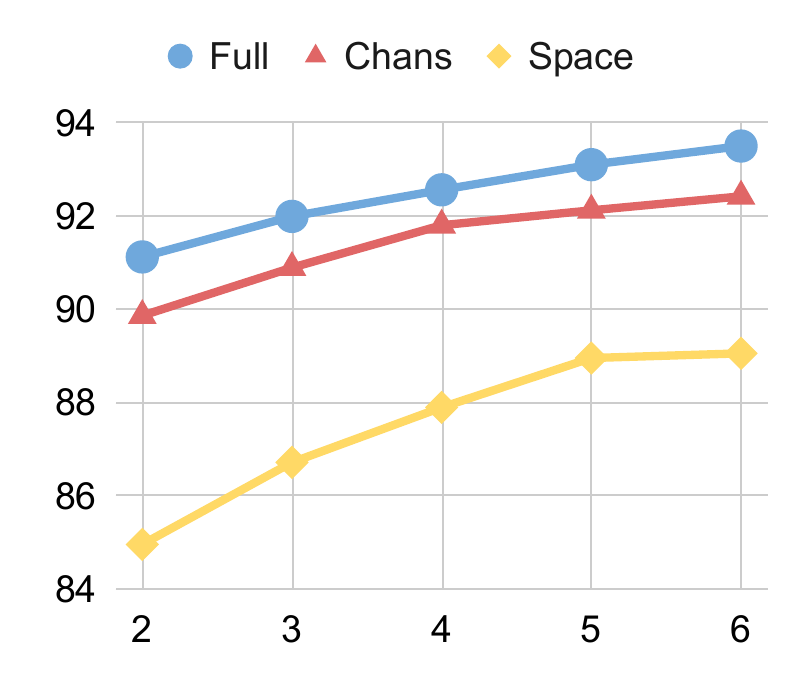} & \rotatebox[origin=c]{90}{Validation Acc. \%} & \hspace{-.5cm} \includegraphics[align=c,width=0.29\linewidth]{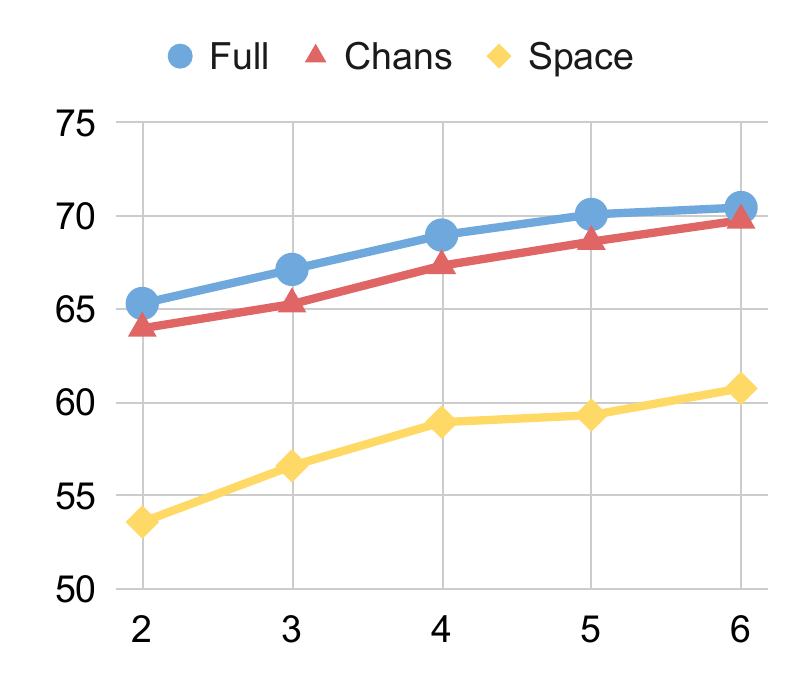} & \rotatebox[origin=c]{90}{Validation Acc. \%} &\hspace{-.5cm} \includegraphics[align=c,width=0.29\linewidth]{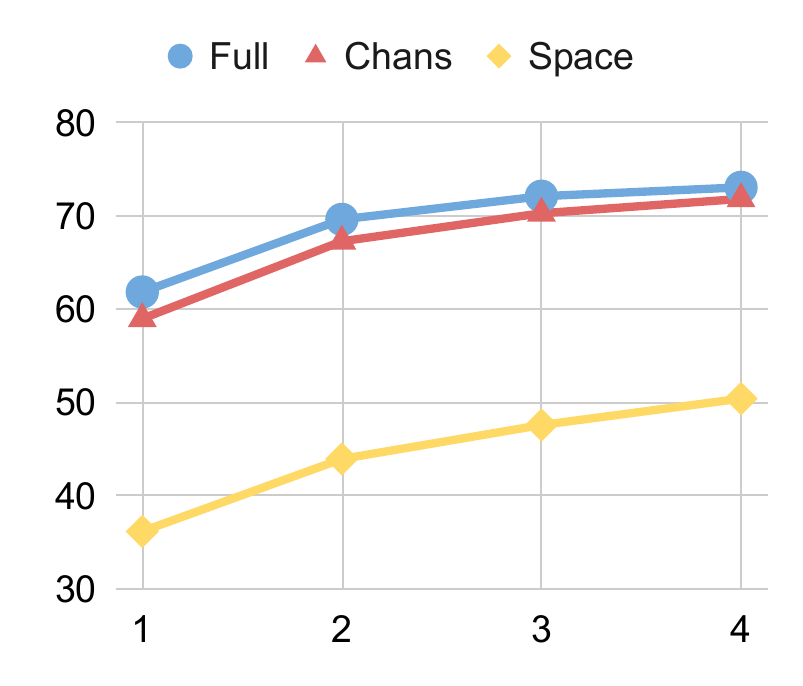} \vspace{-0.1cm}\\
        & Depth && Depth && Depth
    \end{tabular}
    \vspace{-0.25cm}
    \caption{
    \looseness=-1
    While the fully learned ResNets (Full) outperform all others, we see that the models that only learn channel mixing (Chans) remain quite competitive, especially so on ImageNet (right). Conversely, the models that only learn spatial mixing (Space) lag very far behind the others.}
    \vspace{-0.2cm}
    \label{fig:res-acc}
\end{figure*}
\subsection{The ConvMixer}
As one of the latest in an ever-growing line of isotropic vision models \cite{imagegpt,mlpmixer,visiontransformer,yu2022metaformer}, the ConvMixer \cite{convmixer} architecture adapts the (at this point) classical method of convolutions to the equi-sized representation (isotropic) paradigm of transformer models. Perhaps more importantly for our analysis, it serves as a state-of-the-art convolutional model in which \textit{the convolutions are already separable}. In other words, the ConvMixer consists of strictly depthwise and pointwise convolutions. This allows us to examine the effects of learning only channel mixing in a state-of-the-art architecture without making \textit{any} changes to the underlying architecture.

The ConvMixer \cite{convmixer} architecture consists of depthwise convolutions (wrapped in a residual connection) and pointwise convolutions. Specifically, a depth-$n$ ConvMixer consists of a one-layer patch-projection stem, $n$ depthwise-pointwise blocks, a global average pool, and a linear classifier. All throughout, the representation maintains the size to which it is projected by the stem (hence the term \textit{isotropic}).

\subsection{Is Channel Mixing (Almost) All You Need?}
\looseness=-1
By isolating the spatial and channel-mixing parts of the standard convolution into separable components, we can analyze the contribution of channel mixing alone to the success of convolutional neural networks. To do this, we now introduce a form of the separable convolution (illustrated in supplement) wherein we leave the depthwise (spatial-mixing) filters frozen at their initialized values and only learn the pointwise (channel-mixing) filters during training. In our results, this is indicated by ``Chans'' (or ``channels-only'') whereas the fully-learned models are indicated by ``Full'' (or ``fully-learned'').

Note that the number of trainable parameters does not depend on the size of the kernel $(k)$ since only the pointwise parameters are learned. Furthermore, we can ensure the number of trainable parameters in the pointwise convolution is equal to that of the corresponding standard Conv2D by setting $d = k^2$:
\begin{equation}
    \begin{split}
        &d = k^2\\
        \implies & c_{in}\cdot c_{out} \cdot k^2 = c_{in}\cdot d \cdot c_{out}\\
        \implies & p_{conv} = p_{point}
    \end{split}
\end{equation}
For ResNet architectures, the standard convolution kernel is of shape 3$\times$3, so we would set $d=3^2=9$ to use the same number of parameters in our channel-learning-only convolution (and $d=3$ to use one third of the parameters).

Here this is simply used as a heuristic to create models with similar parameter counts to the vanilla ResNet architecture. All of our separable ResNet models will thus have a depth-multiplier of 9, regardless of which parameters are being learned.

\section{Experiments}
After motivating with some relevant background, we now begin our study into the capabilities of networks that only learn channel mixing. 
Section \ref{sec:resnet} applies our hypothesis to classical ResNet \cite{resnet} architectures. 
In Section \ref{sec:convmixer}, we move on to the naturally separable ConvMixer \cite{convmixer} architecture to extend our findings to a state-of-the-art isotropic model. 
Then, in Section \ref{sec:robust}, we explore a practical application in the form of architectural adversarial robustness. 
Lastly, in Section \ref{sec:shuffle}, we extend our findings beyond the classification regime to the pixel un-shuffling problem.
Our code will be made public upon publication.

\subsection{ResNet Experiments}\label{sec:resnet}

For these experiments, we train all models under identical conditions: the same number of epochs, same batch size, same learning rate, same decay schedule, etc. Any and all differences (or similarities) in performance are due entirely to the intrinsic properties of the architectures themselves. Experiments are conducted on CIFAR-10 \cite{cifar}, CIFAR-100,  \cite{cifar} and ImageNet-1k \cite{imagenet}.

\subsubsection{Model Architectures}
For our first set of experiments, we employ the ResNet \cite{resnet} architecture with the identity mapping modifications introduced in \cite{resnetv2} and convert all convolutions to the separable variant as described in Section \ref{sec:conv}. 
A depth-$n$ CIFAR \cite{cifar} ResNet contains a 1-layer stem, 2$n$ layers for each of the 3 blocks, and a linear classifier, for a total of $6n+2$ layers. Similarly an \linebreak ImageNet \cite{imagenet} ResNet contains a 1-layer stem, 2$n$ layers for each of the 4 blocks, and a linear classifier, for a total of $8n+2$ layers. The ``depth'' in our plots represents this $n$. All separable ResNets have a depth multiplier of 9.

\subsubsection{Classification Gap}
\looseness=-1
For our ResNets that learn only channel mixing (Chans), any spatial mixing is done using depthwise filters fixed at their random initializations. Conversely, for those that learn only spatial mixing (Space), any channel mixing is done through the pointwise 1x1 convolutions frozen at their initial states.
\looseness=-1
When we look at the performance of fully-learned ResNets versus those learning \textit{only} channel mixing (Chans), we make an interesting discovery: the gap in classification accuracy is actually quite small (Figure \ref{fig:res-acc}). One might question the generalization of this observation, noting that CIFAR \cite{cifar} is a simple dataset, and the performance may just be saturated. Yet, the trend also holds for ImageNet \cite{imagenet}, a \textit{much} larger and more complex dataset. Furthermore, we also see that networks only learning spatial mixing (Space) consistently perform \textit{far} worse than those learning only channel mixing. 

One might note that the number of spatial mixing parameters in a separable convolution, learned or not, is inherently smaller than that of channel mixing parameters. This is true; however, the same can also be said of the standard convolution. In a typical network, the channel mixing dimensions that contribute to the weight size ($c_{in}\cdot c_{out}$) are typically much larger than the spatial mixing dimensions ($k\cdot k$).

Ultimately, the shown fact that networks learning only channel mixing achieve competitive results to those that are fully learned comes as quite the surprise and calls into question the significance of learning the spatial versus channel-mixing operations in general.

\subsubsection{Static Filter Structure}
\begin{figure}[t]
    \vspace{0cm}
    \centering
        \begin{tabular}{cc}
    & \footnotesize{Depth-2 Separable ResNet (CIFAR-100)} \\
         \rotatebox[origin=c]{90}{\footnotesize{Validation Acc. \%}} & \hspace{-.2cm} \includegraphics[trim={0.4cm 0.4cm 0.8cm 0.4cm},clip,align=c,width=0.8\linewidth]{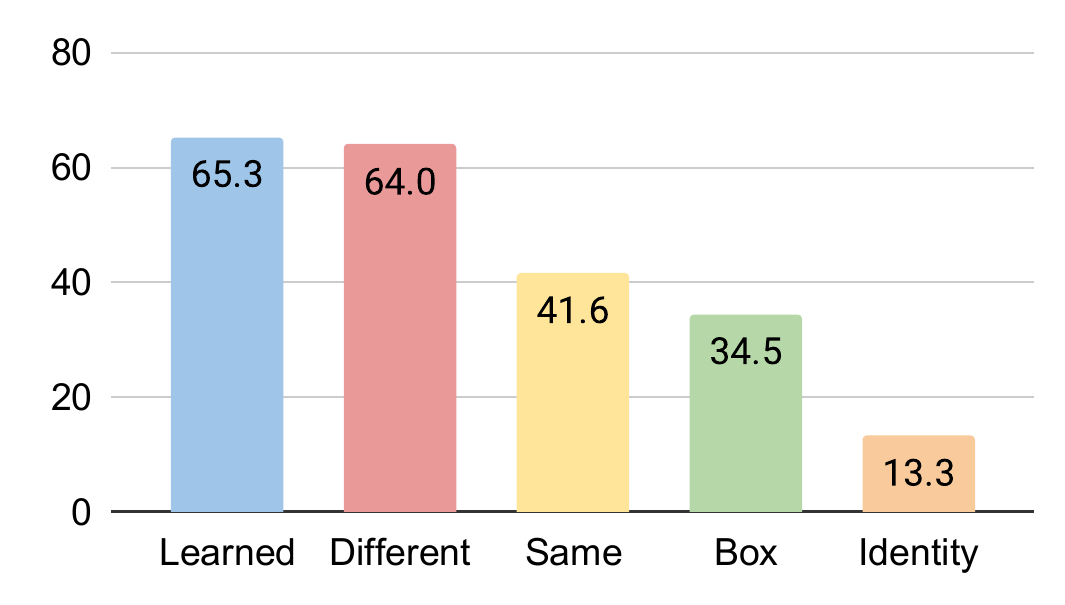}
    \end{tabular}
    \vspace{-0.25cm}
    \caption{While \textit{randomly} initialized filters can provide competitive results, the same is not true for \textit{any} arbitrary, fixed filter. Random filters work best when they are uncorrelated from eachother, allowing them to extract different information.}
    \label{fig:inits}
    \vspace{-0.2cm}
\end{figure}

While we have shown that learning spatial mixing is not necessary to achieve competitive performance without changes to the model's architecture, the static filters cannot be \textit{completely} arbitrary; they must still hold some basic properties to adequately transform the input signal.

Looking at Figure \ref{fig:inits}, we see the performance of the fully-learned separable convolutions as the leftmost (blue) column.  The next column (red) represents the separable convolution that only learns channel mixing. We again note that the randomly initialized spatial mixing performs only slightly worse than the fully-learned version. We label this column ``Different'' to contrast with the next (yellow) column, ``Same.'' For this setting, we apply the \textit{same} set of 9 randomly initialized kernels to \textit{every input channel}. The large performance drop here shows us that highly-correlated filters do not make good feature extractors since they cover the same spectrum of signals. The next drop-off to the box filter (green) occurs because box filters are \textit{perfectly} translationally correlated with each other. Lastly, the catastrophic failure of the identity filter (orange) tells us that although we might not need to \textit{learn} it, spatial mixing is still an integral part of discriminative networks. 
\subsubsection{Extra-Wide ResNets}\label{sec:width}
\begin{figure}[t]
    \centering
    \setlength{\tabcolsep}{1pt}
    \begin{tabular}{cccc}
    & \footnotesize{CIFAR-10} && \footnotesize{CIFAR-100} \\
        \rotatebox[origin=c]{90}{\footnotesize{Validation Acc.}} & \includegraphics[trim={0.4cm 0.4cm 0.4cm 0.4cm},clip,align=c,width=0.45\linewidth]{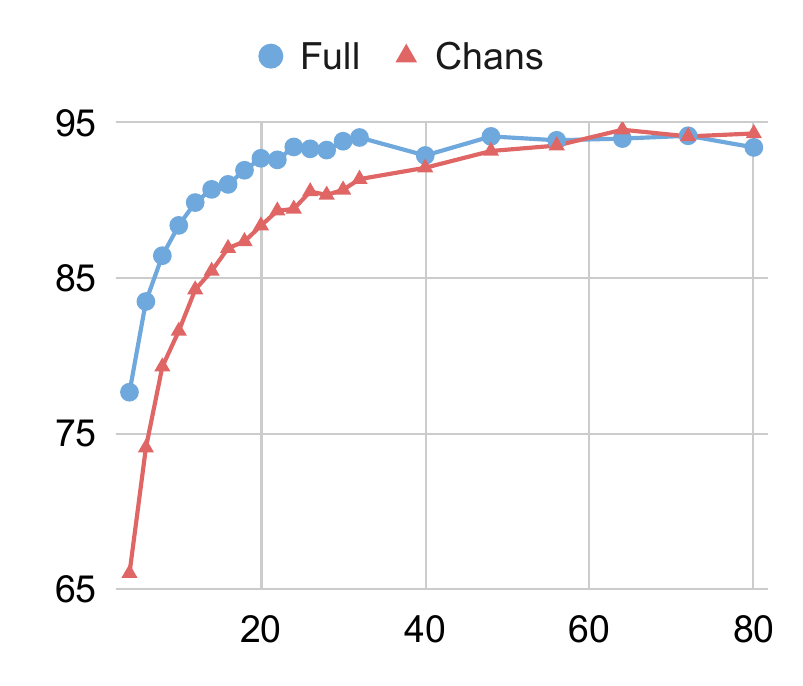}& \rotatebox[origin=c]{90}{\footnotesize{Validation Acc.}} & \includegraphics[trim={0.4cm 0.4cm 0.4cm 0.4cm},clip,align=c,width=0.45\linewidth]{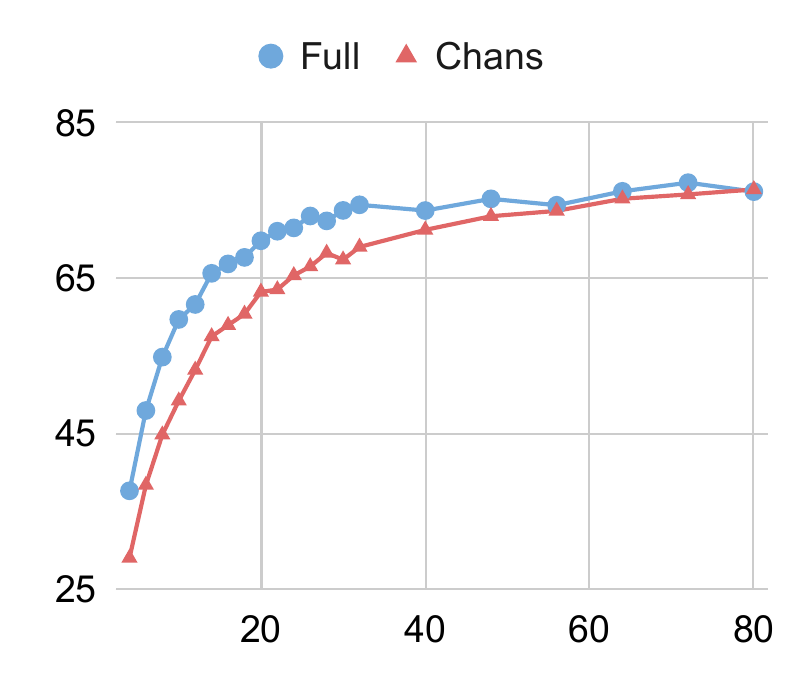}\\
        & \footnotesize{Base Width} && \footnotesize{Base Width}
    \end{tabular}
    \vspace{-0.25cm}
    \caption{
    \looseness=-1
    As we increase the network width, we see the performance of the channels-only models converges to that of the fully-learned models}
    \label{fig:width}
\end{figure}
\looseness=-1
Our previous experiments revealed a trend of the channels-only models only lagging slightly behind the fully-learned models over a range of depths. Now, we examine the gap in performance as we greatly increase the width (number of channels) of the models.

\begin{figure*}[t]
    \centering
    \small
    \begin{tabular}{cccccc}
    &&&\large{ConvMixer}\\
    & CIFAR-10 && CIFAR-100 && ImageNet\\
        \rotatebox[origin=c]{90}{Validation Acc. \%} &\hspace{-.5cm} \includegraphics[align=c,width=0.29\linewidth]{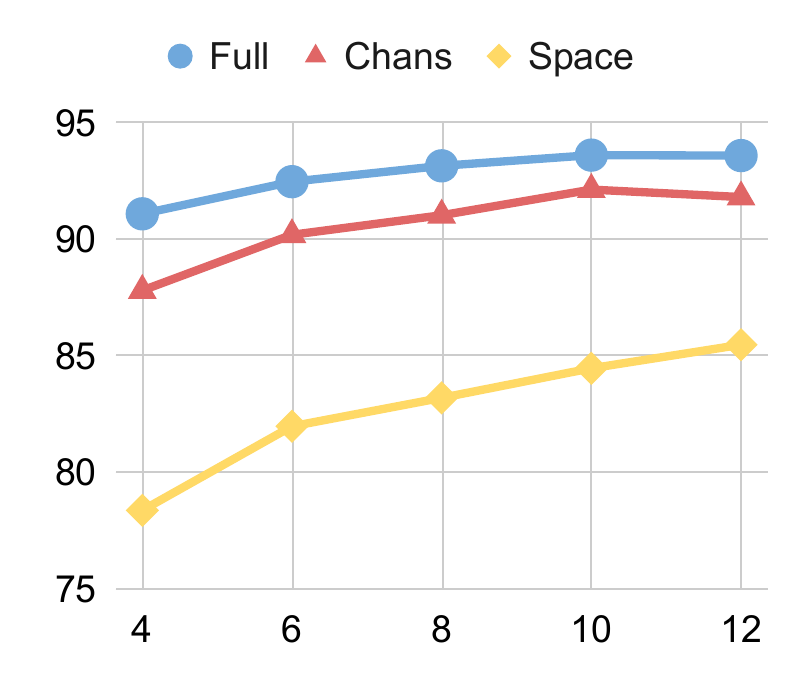} & \rotatebox[origin=c]{90}{Validation Acc. \%} & \hspace{-.5cm} \includegraphics[align=c,width=0.29\linewidth]{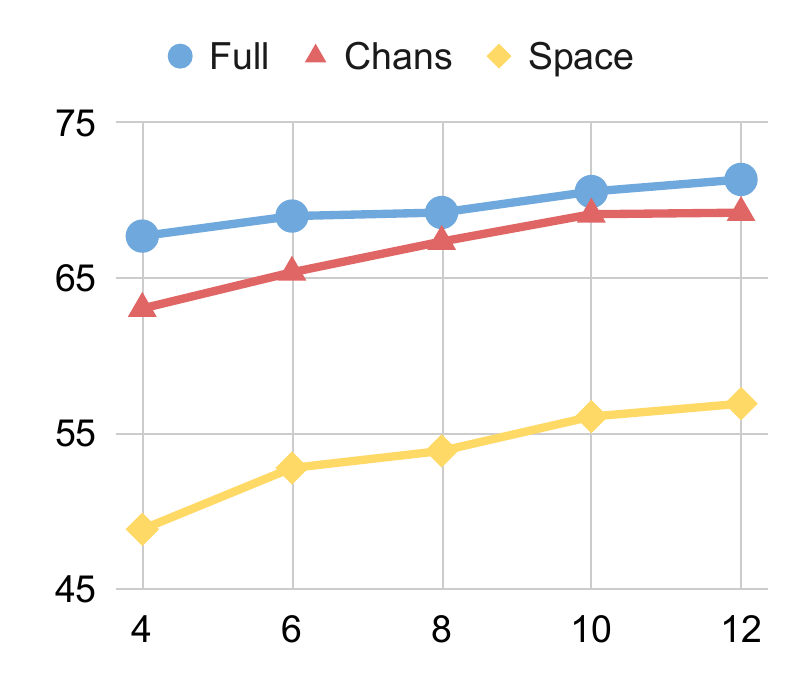} & \rotatebox[origin=c]{90}{Validation Acc. \%} &\hspace{-.5cm} \includegraphics[align=c,width=0.29\linewidth]{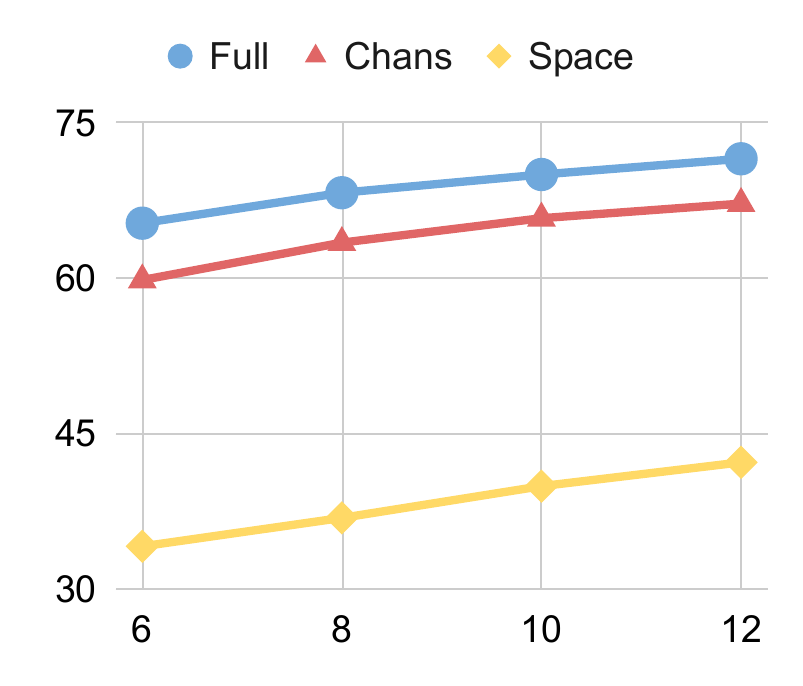} \vspace{-0.1cm}\\
        & Depth && Depth && Depth
    \end{tabular}
    \vspace{-0.25cm}
    \caption{With the ConvMixer architecture, we can better analyze the direct contributions of spatial and channel mixing without altering the original model. We again see networks that only learn channel mixing (Chans) remain competitive with their fully-learned counterparts (Full) while completely out-classing those that only learn spatial mixing (Space).}
    \vspace{-0.3cm}
    \label{fig:conv-acc}
\end{figure*}
\looseness=-1
In Figure \ref{fig:width}, we see that the performance of the channels-only models converges to that of the fully-learned models as we continue to increase the width. While the lottery ticket hypothesis \cite{frankle2018lottery} states that there typically exist optimal randomly initialized sub-networks that \textit{can be trained} to reach the same performance as the full model, our results here suggest that given enough spatial-mixing filters (or ``tickets''), a channels-only model can learn to exploit the filters to \textbf{reach the} \textbf{same performance} as the fully-learned model while \textbf{leaving the spatial-mixing filters at their random initialization}.
\subsection{ConvMixer Experiments}\label{sec:convmixer}

We use the code provided along with the ConvMixer \cite{convmixer} paper for these experiments and employ models with 128 channels, a kernel size of 8, and a patch size of 1 for our CIFAR \cite{cifar} experiments and models with 512 channels, a kernel size of 8, and a patch size of 7 for our ImageNet \cite{imagenet} experiments. Note that in channels-only models, the patch-projection layer is also left at its random initialization.

\begin{figure*}[t]
    \centering
    \begin{tabular}{cccccc}
    &&&\large{Attacked ResNets}\\
    \small
    & FGSM && PGD-2 && FGSM-Defended\vspace{-.1cm}\\
        \rotatebox[origin=c]{90}{Adversarial Acc. \%} &\hspace{-.5cm} \includegraphics[align=c,width=0.29\linewidth]{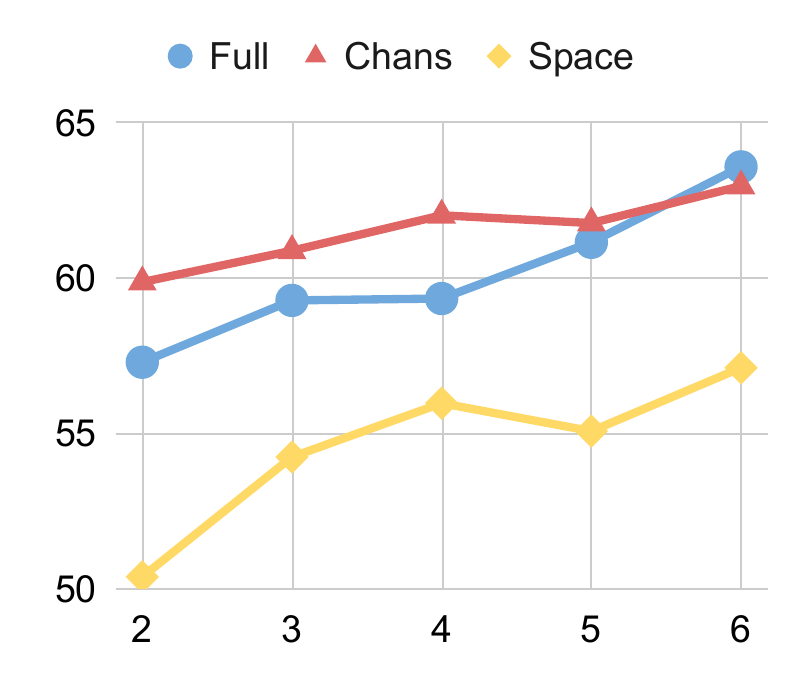} & \rotatebox[origin=c]{90}{Adversarial Acc. \%} & \hspace{-.5cm} \includegraphics[align=c,width=0.29\linewidth]{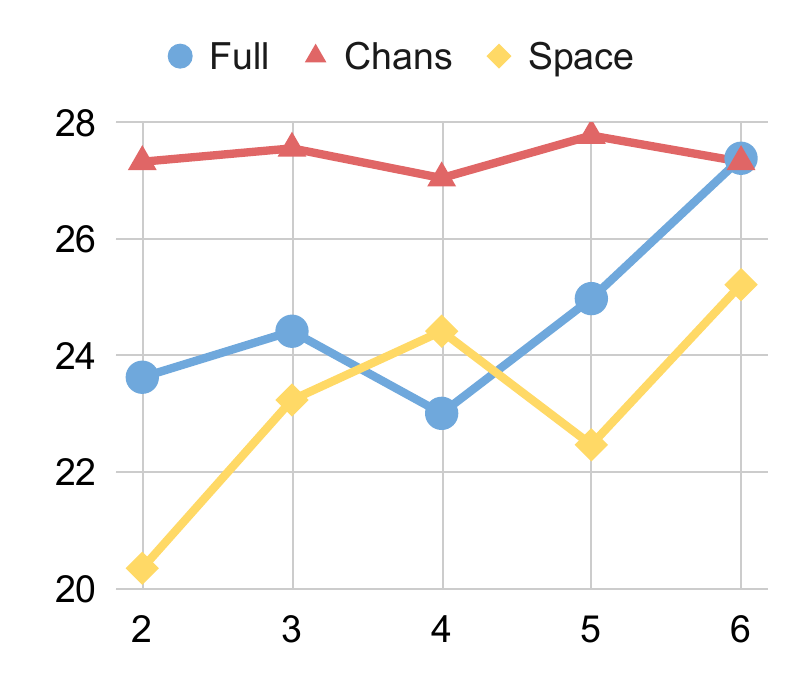} & \rotatebox[origin=c]{90}{Adversarial Acc. \%} &\hspace{-.5cm} \includegraphics[align=c,width=0.29\linewidth]{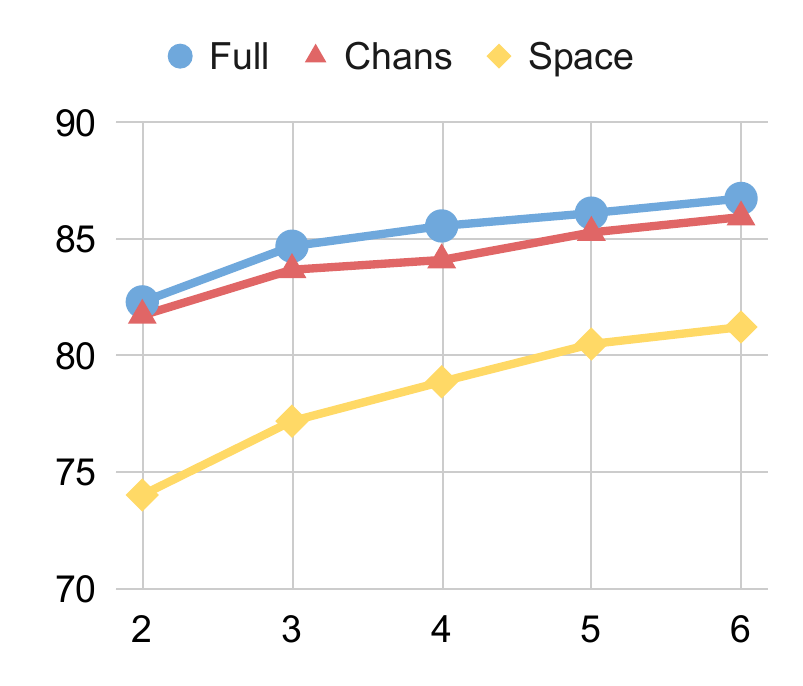} \vspace{-0.0cm}\\
        & Depth && Depth && Depth
    \end{tabular}
    \vspace{-0.25cm}
    \caption{Adversarial accuracy in separable ResNets (CIFAR-10). Channels-only models (Chans) are clearly more \textit{naturally} robust to adversarial perturbations than their fully-learned counterparts and those that learn only spatial mixing for both simple (left) and more advanced (middle) attacks. However, this gap disappears when adversarial training techniques are used (right). ($\epsilon = 1/255$)}
    \label{fig:resadv}
\end{figure*}

\begin{figure*}[t]
\setlength{\tabcolsep}{2pt}
    \centering
    \vspace{-0.5cm}
    \begin{tabular}{ccccc}
         \rotatebox[origin=c]{90}{Rel. Acc. Improvement} &\hspace{-.1cm} \includegraphics[trim={0.8cm 0.4cm 0.2cm 0.4cm},clip,align=c,width=0.28\linewidth]{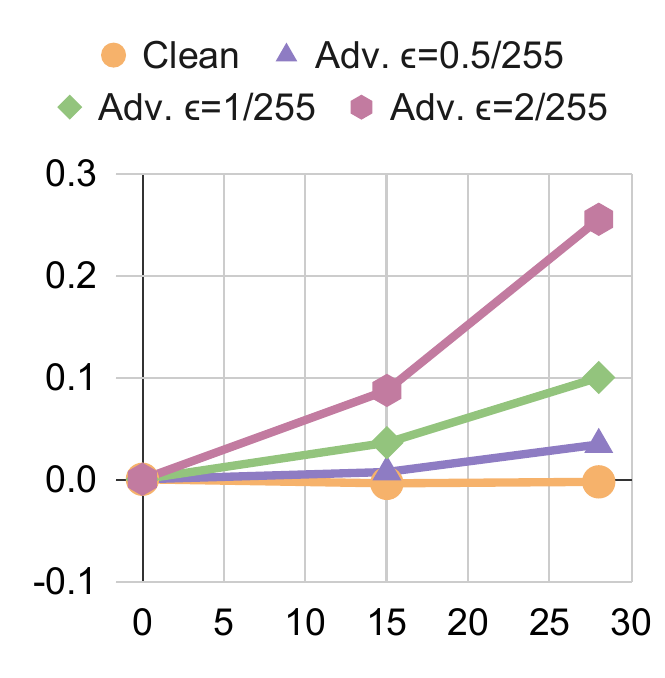} & \includegraphics[align=c,width=0.22\linewidth]{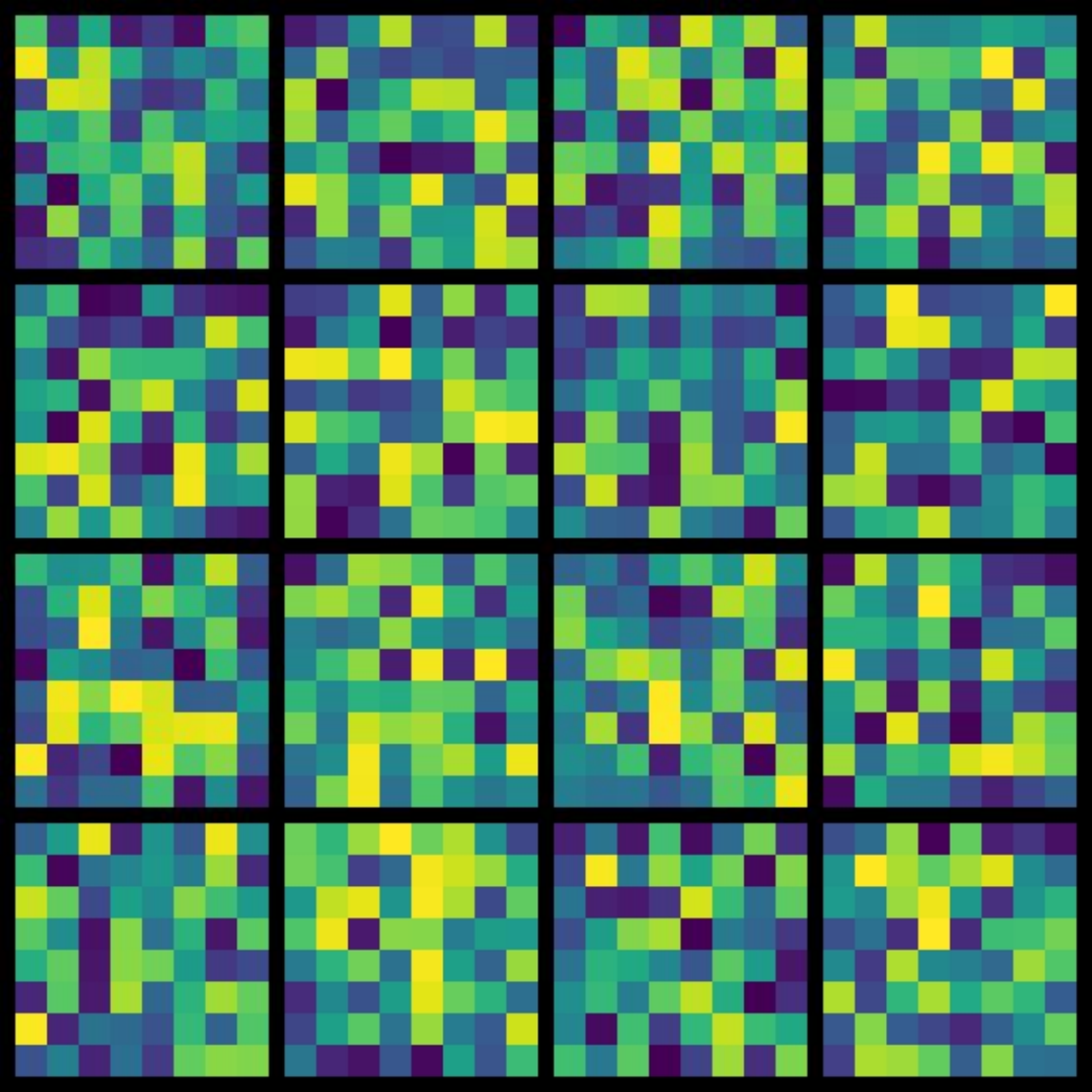}
         & \includegraphics[align=c,width=0.22\linewidth]{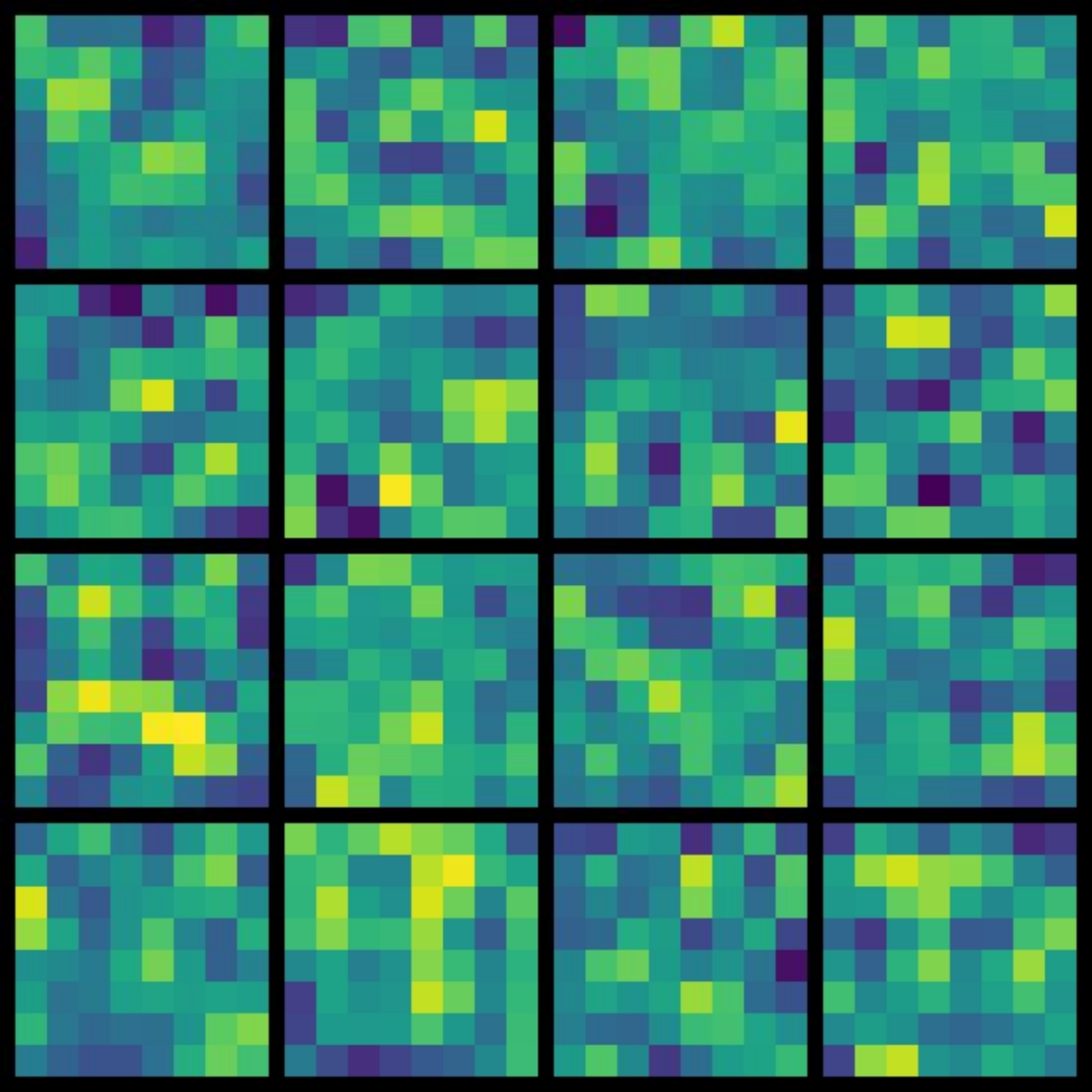}
         & \includegraphics[align=c,width=0.22\linewidth]{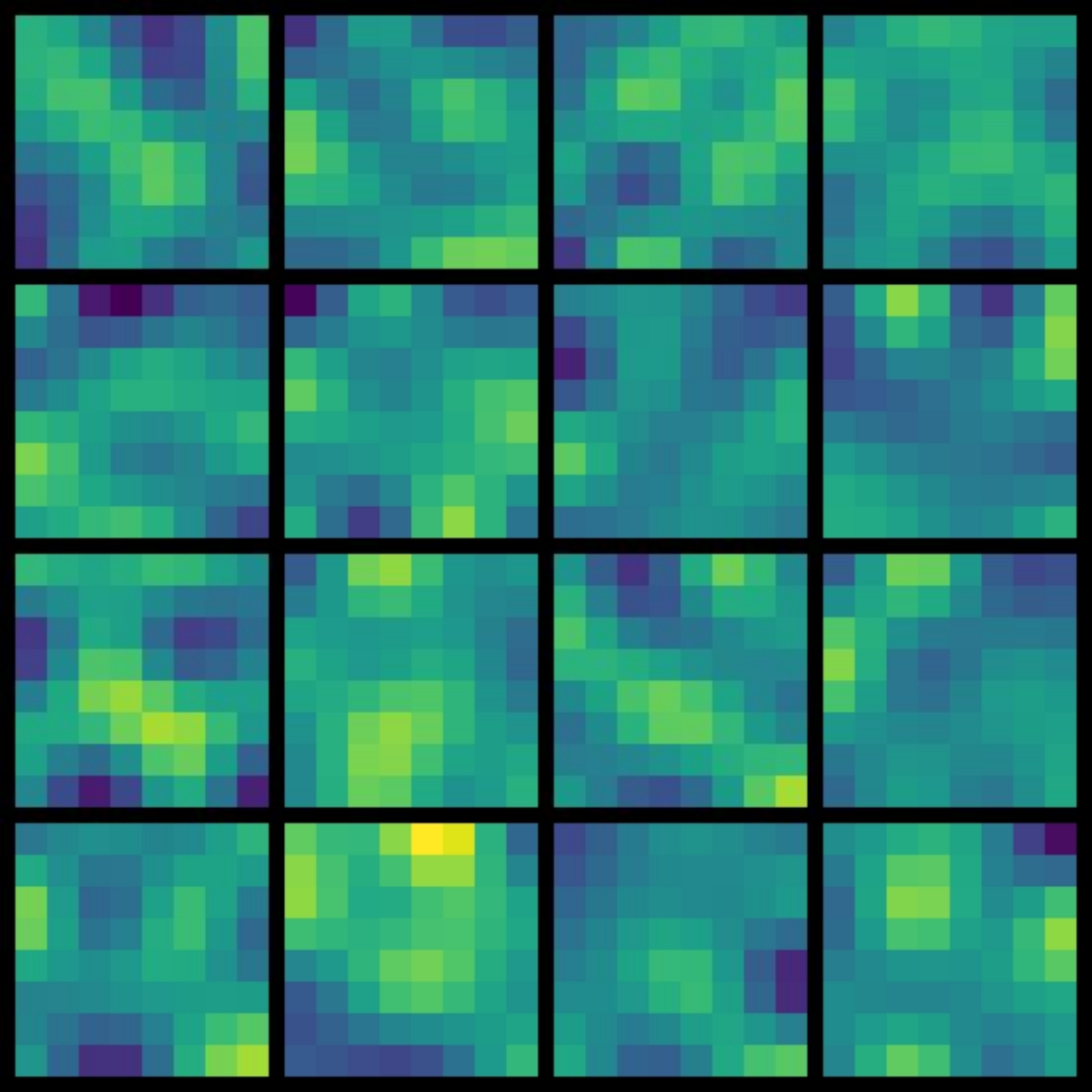}\\
         & Smoothing Index & Random Filters & Smoothed & More-Smoothed
    \end{tabular}
    \vspace{-0.3cm}
    \caption{Smoothing the random filters causes our trained models become significantly more robust to adversarial perturbations.}
    \vspace{-0.2cm}
    \label{fig:smooth}
\end{figure*}

\subsubsection{Naturally Separable Performance}
\looseness=-1
Now that we are dealing with a naturally separable architecture, all comparisons can be made \textit{directly} with the original model. In Figure \ref{fig:conv-acc}, we again see that if we learn only channel mixing, we still achieve results competitive with the fully-learned architecture. Even on the much more challenging ImageNet dataset, our channels-only ConvMixers only lag slightly behind the full-learned models at all depths, showing that this phenomenon extends past toy models and small datasets to state-of-the-art architectures and real-world imagery.

\subsubsection{Filter Observations}

After years of networks using mostly 3$\times$3 filters, the \linebreak ConvMixer \cite{convmixer} takes a step back to larger filters that are more interpretable to the human eye. In Figure \ref{fig:convmixercoverage}, we see the learned spectral coverage of a depth-4 ConvMixer and note that the filters of each layer seem to follow a different distribution. Despite the structure clearly present in these learned filters, the random filters (as seen in Figure \ref{fig:smooth}) still yield comparable performance. As we see in the following section, these specialized learned filters bring with them their own issues in the form of sensitivity to adversarial noise.

\subsection{Adversarial Robustness}\label{sec:robust}
\looseness=-1
Another practical application of our discovery lies in the adversarial setting. Adversarial examples are those specifically curated to fool a neural network into making a mistake at inference time by adding targeted noise to the sample \cite{adversarial}. Typically, the adversarial perturbations are small enough in magnitude that the semantic meaning of the example does not change and that a human observer cannot even notice them.

\subsubsection{The Fast Gradient Sign Method}
\looseness=-1
First, we will quickly introduce one of the first and most simple adversarial attacks developed: the Fast Gradient Sign Method \cite{adversarial}. This method calculates the adversarial perturbations by taking the gradient of the network's loss with respect to the target image and mapping them to the $\ell_{\infty}$ ball. 

Given a neural network ($\mathcal{F}$), target sample ($\mathbf{x}$), and label ($y$), the adversarial example ($\tilde{\mathbf{x}}$) is calculated as
\begin{equation}\label{eq:fgsm}
    \tilde{\mathbf{x}} \coloneqq \mathbf{x} +\; \epsilon \cdot \mbox{sign}\left(\frac{\partial\mathcal{F}(\mathbf{x}, y)}{\partial \mathbf{x}}\right)
\end{equation}
where $\epsilon$ is the radius of the $\ell_\infty$ ball.

This is a simple attack that can be defended against by a variety of methods \cite{cazenavette2021architectural,adversarial,romano2020adversarial,li2019adversarial,chakraborty2018adversarial}, but it is still a useful tool for analyzing the natural adversarial robustness of an architecture without taking any preventative measures.

We also experiment with a more difficult version of this attack where projected gradient descent  (PGD) is used to iteratively update the adversarial example using Eq. \ref{eq:fgsm}. Many updates caused the attack to be too strong for good comparisons, so we just used 2 iterations (PGD-2).

For the experiments where we also employed a defensive technique, FGSM was simply applied to the training samples during model training time.
\subsubsection{Natural Robustness in Separable ResNets}
As we see in Figure \ref{fig:resadv}, ResNets with separable convolutions that only learn channel mixing (Sep-Res) are significantly more robust to the adversarial perturbations than the fully-learned separable ResNets. Without any further modification to the training process, the models that only learn channel mixing seem to be less susceptible to targeted noise.

\begin{figure}
    \centering
    \includegraphics[width=\linewidth]{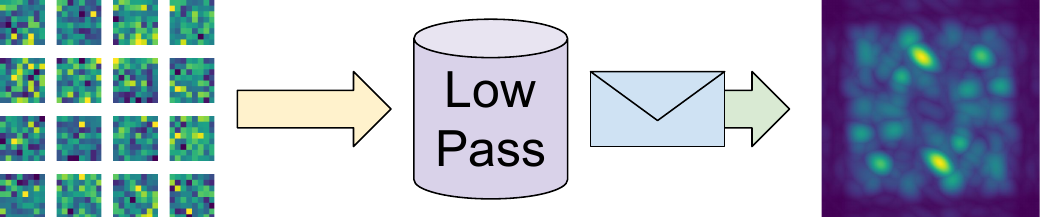}
    \vspace{-0.5cm}
    \caption{By filtering out the high-frequency components of our filter bank, we also eliminate the high-frequency parts of its spectral envelope, leading to a more adversarially robust model.}
    \vspace{-0.3cm}
    \label{fig:low-pass}
\end{figure}
\begin{figure}
    \centering
    \setlength{\tabcolsep}{1pt}
    \begin{tabular}{c c c c}
       \includegraphics[width=.24\linewidth]{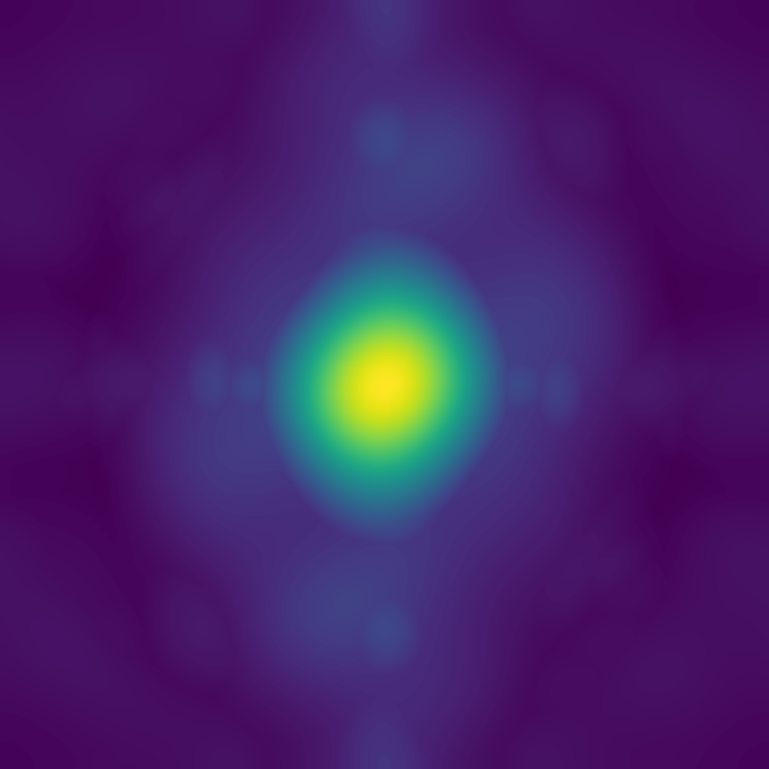}  &  \includegraphics[width=.24\linewidth]{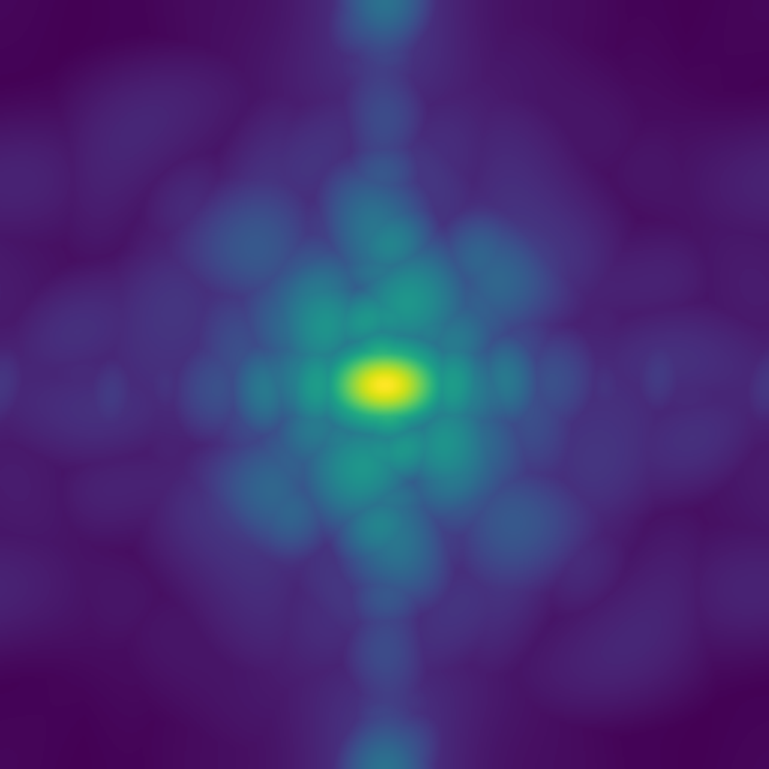}& \includegraphics[width=.24\linewidth]{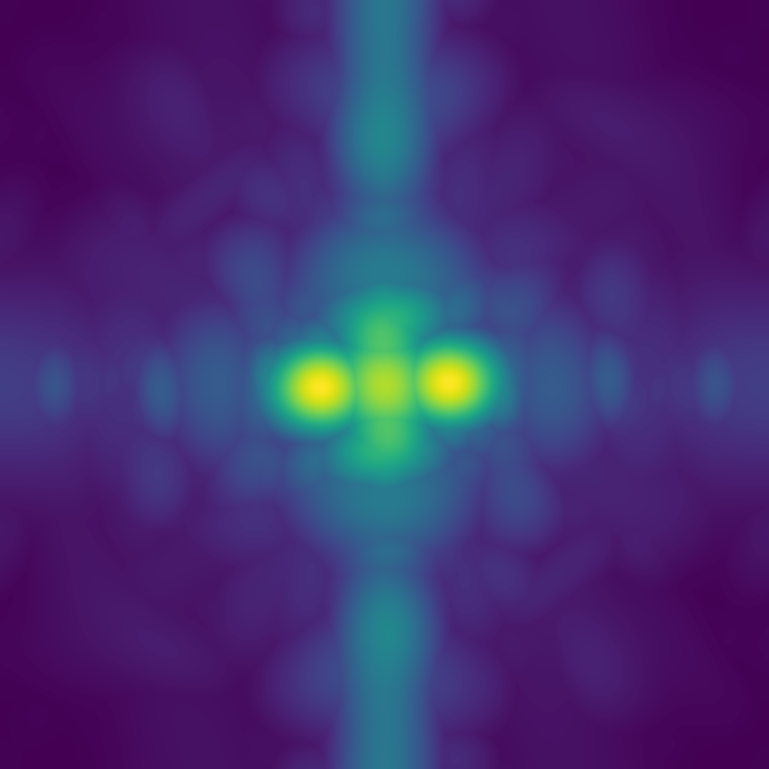} &  \includegraphics[width=.24\linewidth]{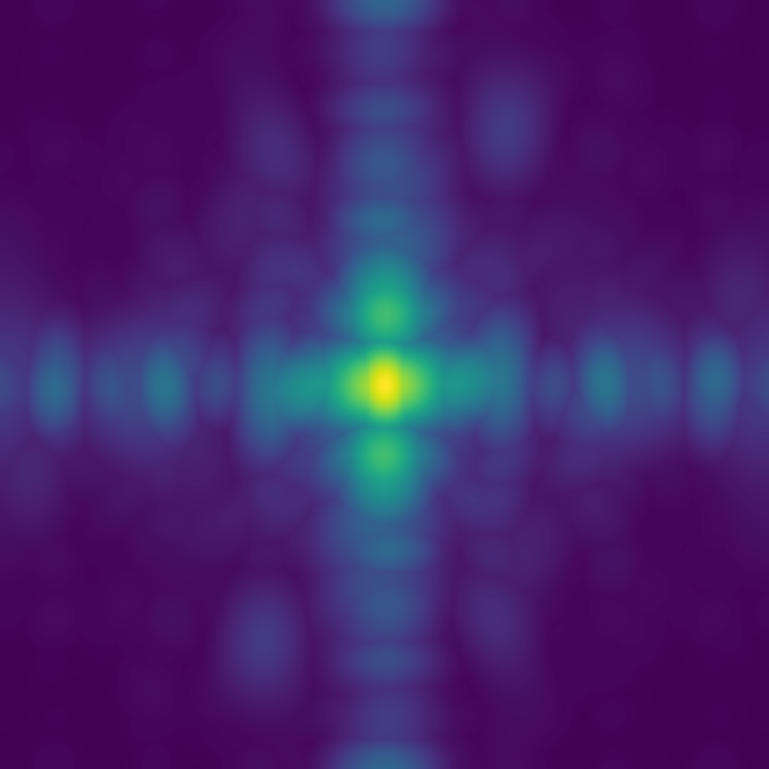}\\
       Layer 1  & Layer 2 & Layer 3  & Layer 4
    \end{tabular}
    \vspace{-0.3cm}
    \caption{Spectral envelopes of learned ConvMixer filters. Naturally-learned filters have dense coverage of the lower frequencies without unnecessarily capturing the high frequencies.}
    \label{fig:convmixercoverage}
\end{figure}
\subsubsection{Induced Robustness in ConvMixers}

\looseness=-1
Looking at envelopes of the learned filters (Figure~\ref{fig:convmixercoverage}) , one could mistakenly link the \textit{smoothness} of learned filters to their poor adversarial robustness, relative to that of random filters (Figure \ref{fig:resadv}). Our next experiment disproves this.

\looseness=-1
We can artificially increase the smoothness of our \textit{random} filters by applying a low-pass smoothing operation directly to the filters themselves. After doing so, the high-frequency components of our random filters have been removed, and the filters now focus more on the mid to low frequencies, just as the learned filters do, as seen in Figure \ref{fig:low-pass}.

\begin{figure*}[h!]
    \vspace{-0.5cm}
    \centering
    \begin{tabular}{cccccc}
    & \scriptsize{$k=3$} && \scriptsize{$k=5$} && \scriptsize{$k=7$} \vspace{-0.1cm}\\
        \rotatebox[origin=c]{90}{Test PSNR~~~~~~~~} &\hspace{-.1cm} \includegraphics[trim={0.8cm 0.4cm 0.2cm 0.4cm},clip,align=c,width=0.25\linewidth]{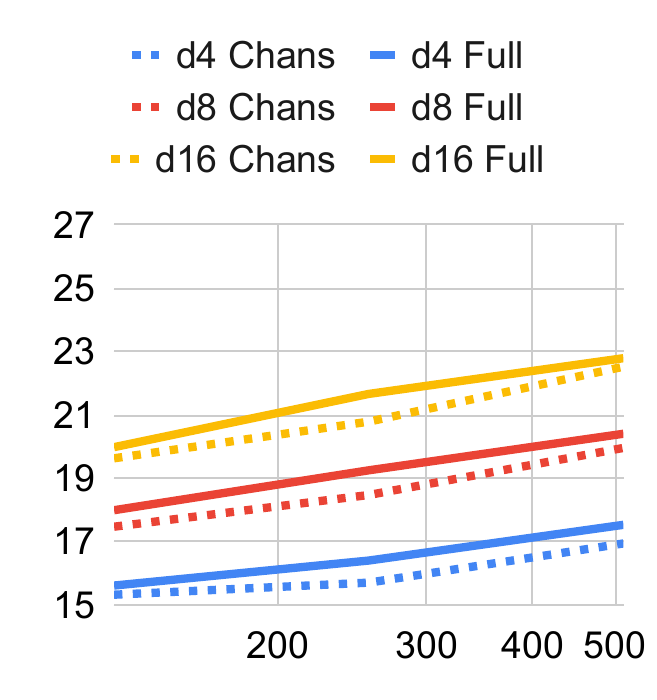} & \rotatebox[origin=c]{90}{Test PSNR~~~~~~~~} & \hspace{-.1cm} \includegraphics[trim={0.8cm 0.4cm 0.2cm 0.4cm},clip,align=c,width=0.25\linewidth]{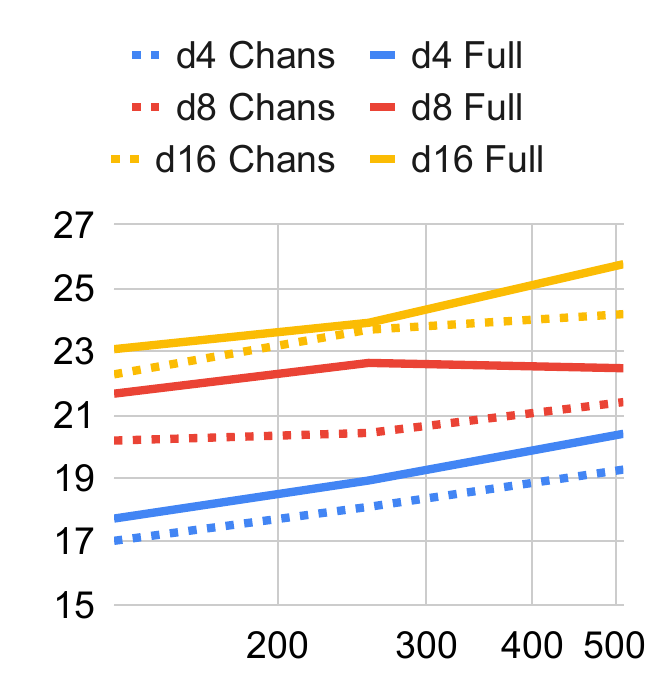} & \rotatebox[origin=c]{90}{Test PSNR~~~~~~~~} &\hspace{-.1cm} \includegraphics[trim={0.8cm 0.4cm 0.2cm 0.4cm},clip,align=c,width=0.25\linewidth]{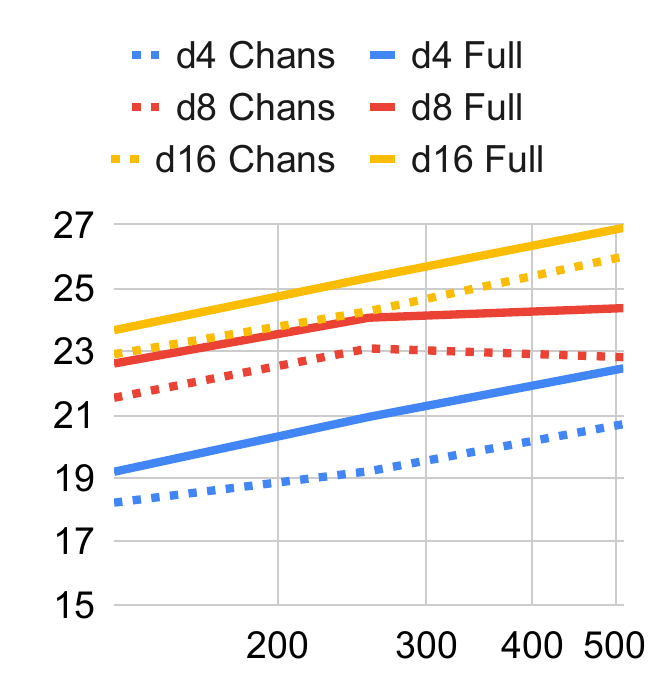} \vspace{-0.1cm}\\
        & Width && Width && Width
    \end{tabular}
    \vspace{-0.25cm}
    \caption{\textbf{CIFAR-10}: PSNR for Pixel Un-Shuffling. Once again, yet for a completely different task, we see that channels-only models only lag slightly behind their fully-learned counterparts over a wide variety of network sizes.}
    \label{fig:pnsr}
\end{figure*}
\looseness=-1
In Figure \ref{fig:smooth}, we see that instead of hurting performance, smoothing the random filters actually significantly \textit{increases} their robustness to adversarial attacks while not hurting performance on clean data (orange line) at all. In fact, we see that smoothing results in a relative performance increase of over 25\% for the highest magnitude of attack. We then conclude that there must be a different reason (statistical shortcuts, etc.) for the learned filters' poor robustness.

\subsection{Pixel Un-Shuffling}\label{sec:shuffle}
\looseness=-1
Thus far, we have only shown the power of channels-only models as they apply to classification tasks.

\begin{figure}[h!]
    \centering
    \footnotesize
    \begin{tabular}{cc}
        \includegraphics[width=0.45\linewidth]{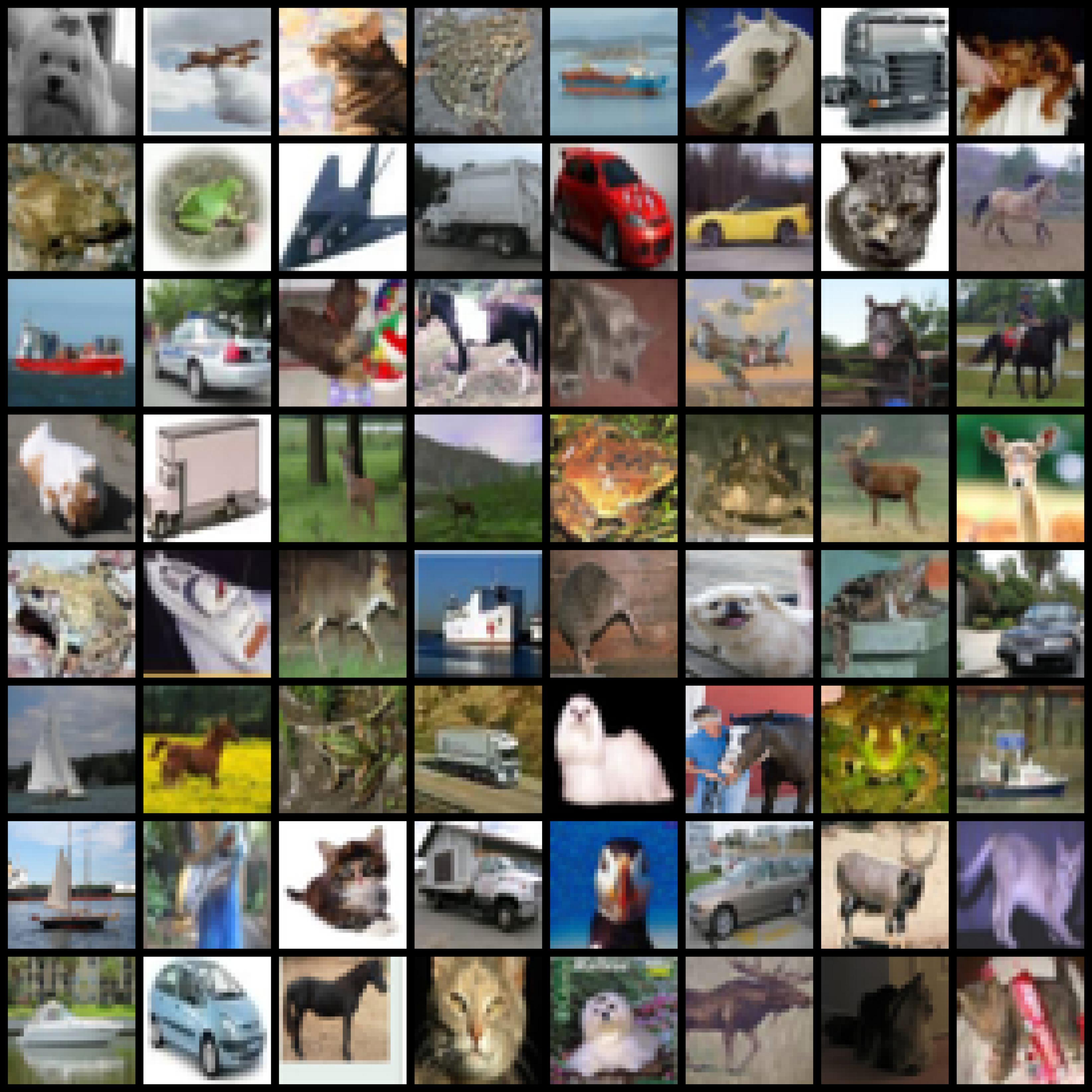} & \includegraphics[width=0.45\linewidth]{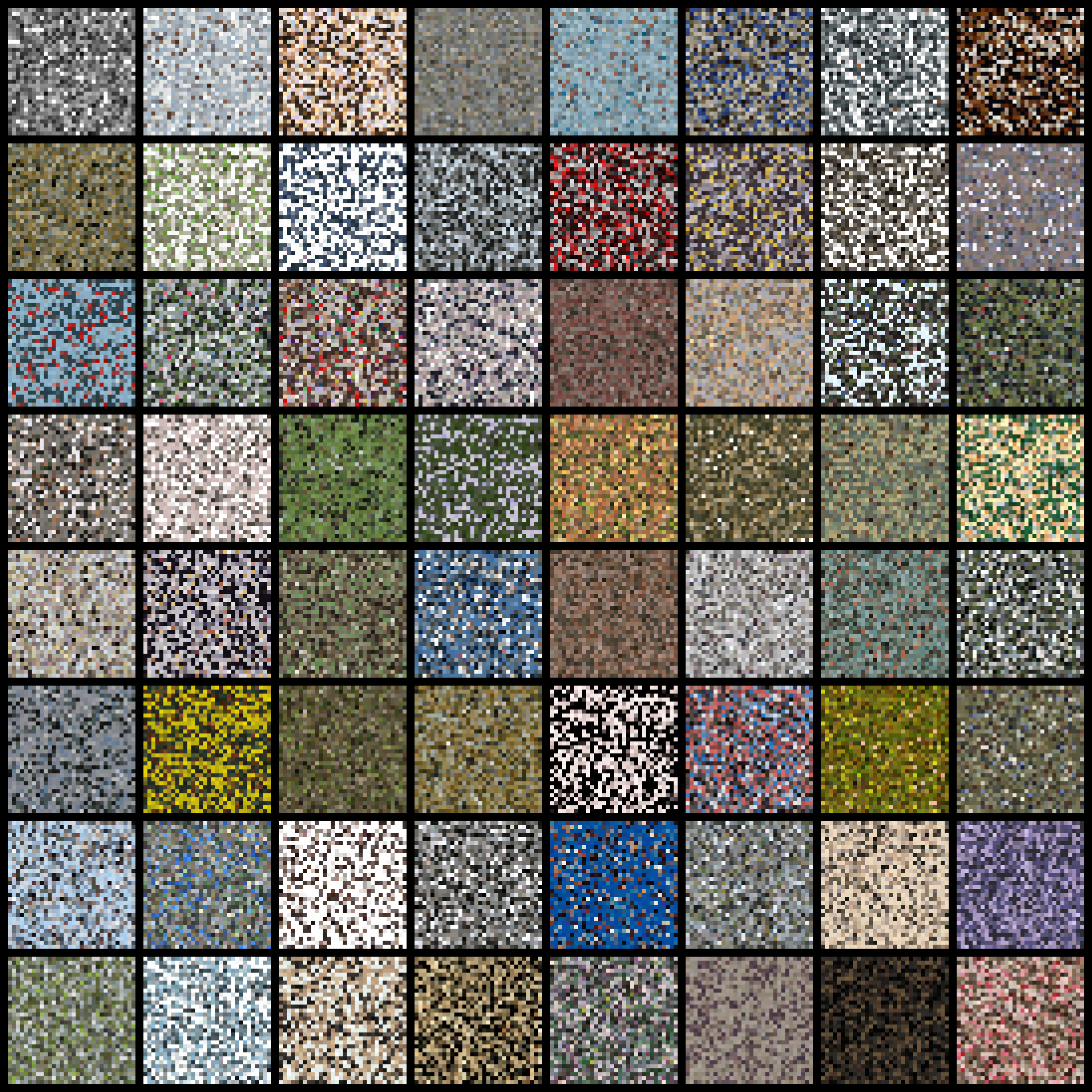} \\
        (A) Ground Truth & (B) Permuted Pixels \\
        \includegraphics[width=0.45\linewidth]{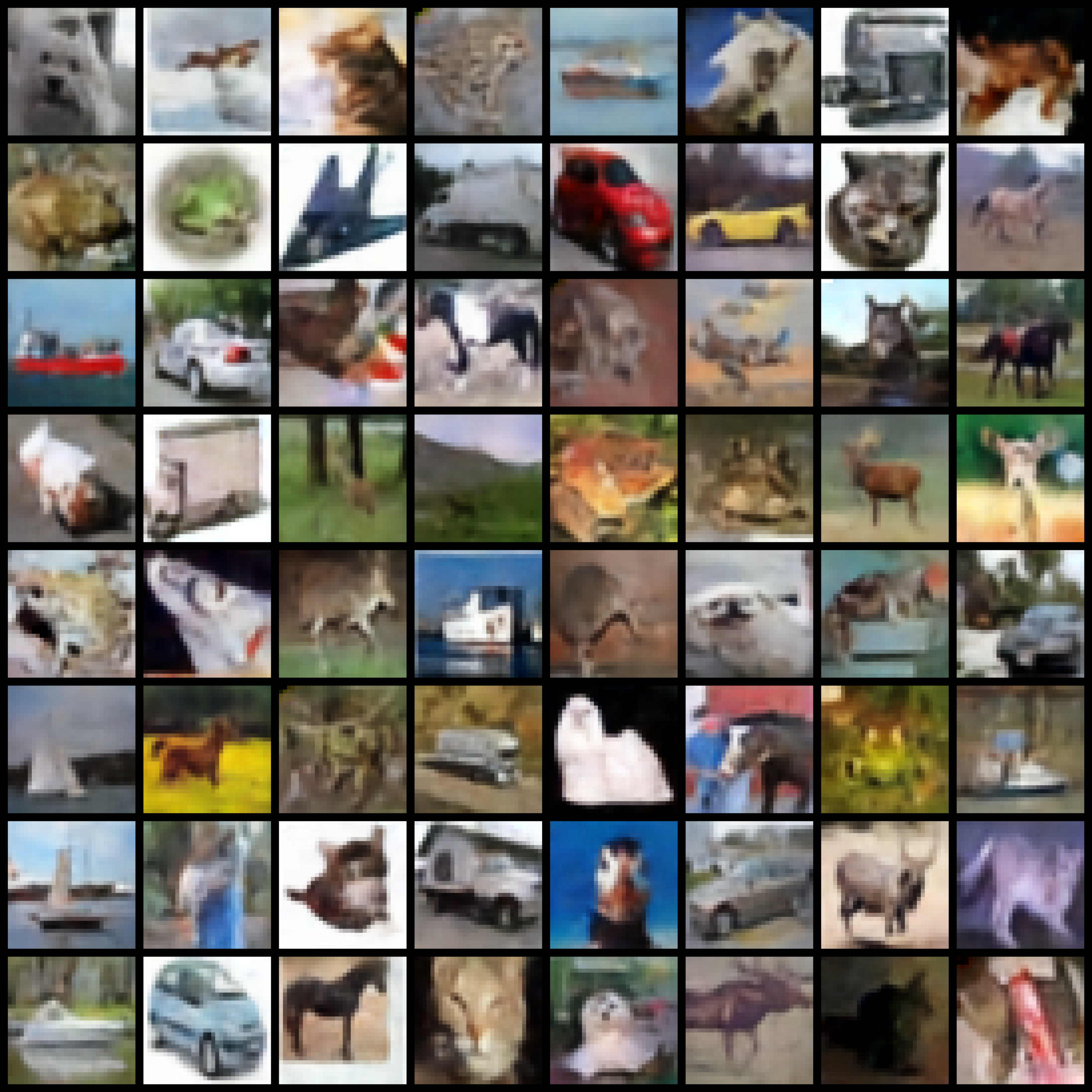} & \includegraphics[width=0.45\linewidth]{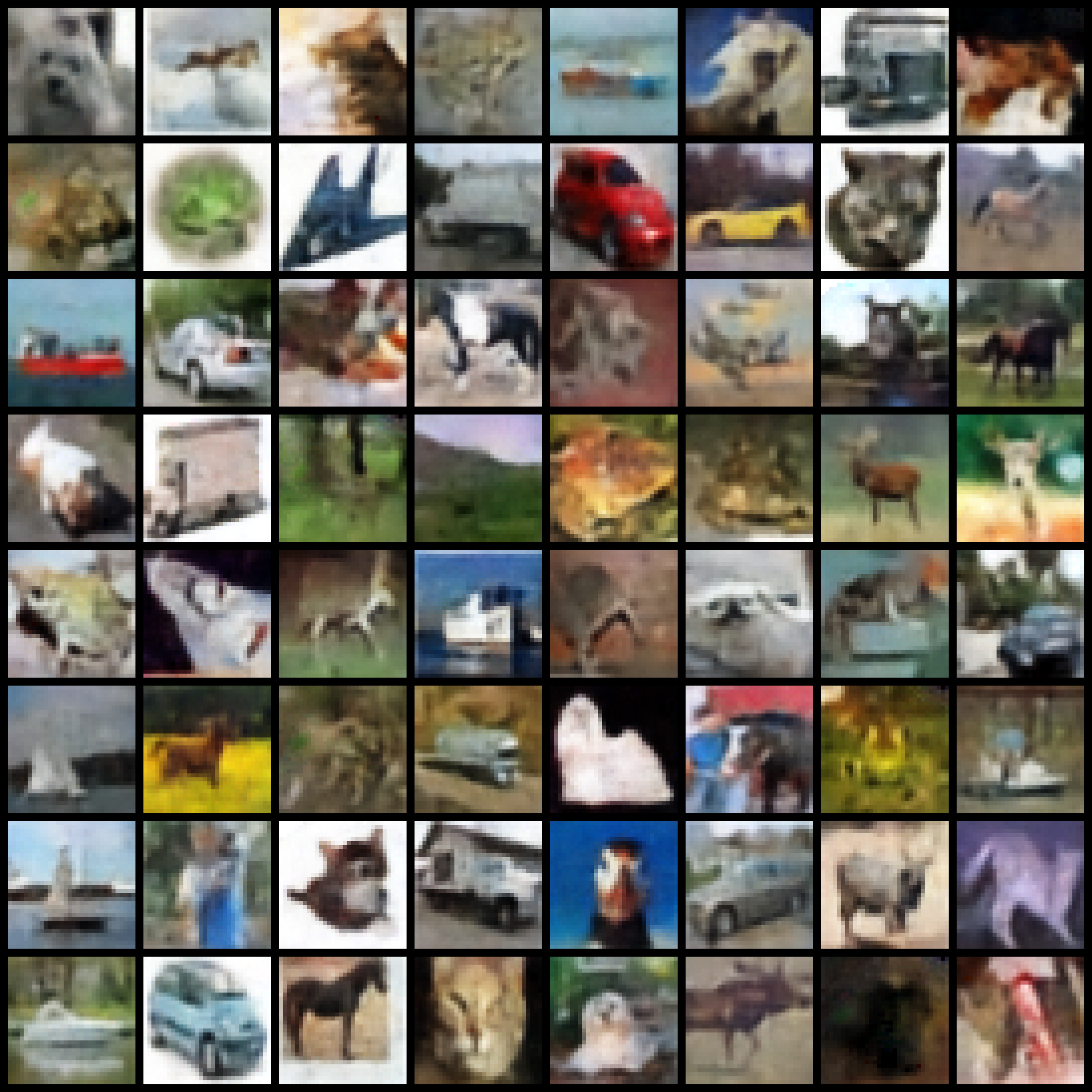} \\
        (C) Large-Model Recon. (Full) & (D) Large-Model Recon. (Chans)\\
    \end{tabular}
    \vspace{-0.3cm}
    \caption{Channels-only un-shuffling models perform comparably to fully-learned models on the pixel un-shuffling task, indicating that the observed phenomena extends past classification.}
    \label{fig:cifar}
    \vspace{-0.5cm}

\end{figure}

As evidence that such models can also prove effective for other tasks, we now apply them to the \textit{pixel un-shuffling} problem. We define this problem as applying a deterministic random permutation of the spatial coordinates of all the pixels of the input image and tasking the model with reconstructing the un-corrupted signal. (To be clear, the same permutation pattern is applied to every input image.)

For images $\mathbf{x}_i$, pixel permutation $\sigma$, and network $\mathcal{F}$ with parameters $\theta$, our formal training objective becomes

$$\min_\theta \frac{1}{n}\sum_{i=1}^n \|\mathbf{x}_i - \mathcal{F}(\sigma(\mathbf{x}_i); \theta)\|^2$$
\looseness=-1
For this task, we use a patch-size 1 ConvMixer with a slight modification: instead of a global average pool and classification head, we simply apply a linear projection back up to the appropriate number of channels (1 or 3). We chose the ConvMixer because its isotropic structure makes it ideal for image-to-image translation, and its natively separable design allows for easy analysis of the effects of spatial and channel mixing. 

\looseness=-1
Intuitively, one would think the learning of spatial mixing is crucial for the task of ``moving'' pixels back to their correct location. However, our experiments show that, yet again, channels-only models perform nearly just as well as their fully-learned counterparts. Randomly selected examples from the test sets can be seen in Figure and \ref{fig:cifar}.

\looseness=-1
On all of MNIST \cite{deng2012mnist}, Fashion-MNIST \cite{xiao2017fashion} and CIFAR-10 \cite{cifar}, the qualitative and quantitative test-set results for both model types are strikingly close. Unlike the classification problem, these results give us a clear, human-interpretable notion of the fact that networks can seemingly learn effective point-wise linear transforms to leverage the natural spatial mixing provided by the randomly-initialized depth-wise convolutions. In Figure \ref{fig:pnsr}, we include plots showing PNSR trends for various model architectures on CIFAR-10. We see that the performance increases with depth, width, and kernel size despite the parameter count of channels-only models not scaling \textit{at all} with kernel size.
(We include results for MNIST and Fashion-MNIST in the supplementary material.)

\section{Conclusion}
While previous works have shown the merits of fully frozen networks as feature extractors \cite{saxe2011random,huang2006extreme}, none (to the best of our knowledge) have explored the question of which parameters are the most important or beneficial to learn. 

In this work, we have done just that; we have shown empirically that networks that only learn channel mixing can reach nearly the same performance as fully-learned networks without any further alterations. For intuition, we offer a possible explanation via the spectral envelopes of the random filter banks reaching dense coverage. We also showed that networks with random spatial-mixing weights are naturally more robust to adversarial attacks, and we also offer a method for \textit{further} increasing the adversarial robustness of such networks by artificially altering their spectral coverage. Lastly, we showed that this phenomenon extends past the classification regime and that such restricted models are also capable of learning other tasks, such as pixel un-shuffling.

\clearpage
{\small
\bibliographystyle{ieeenat_fullname}
\bibliography{main}
}

\ifarxiv \clearpage \appendix
\counterwithin{figure}{section}
\onecolumn

\section{The Separable Convolution}
In Figure~\ref{fig:static}, we show an illustration of the separable convolution. 
Throughout this work, "Full" refers to both the depthwise and pointwise filters being learned. 
"Space" refers to only the depthwise (spatial-mixing) filters being learned. 
Likewise, "Chans" refers to only the pointwise (channel-mixing) filters being learned.

\section{More Pixel Un-Shuffle Results}

For our CIFAR pixel un-shuffling figure in the main text (Figure~\ref{fig:cifar}), the ``large'' model had a depth of 16, width of 512, and kernel size of 7.

On the following pages, we include further results on pixel un-shuffling that were omitted from the main text due to space constraints.

\begin{figure*}[!h]
    \centering
    \includegraphics[width=\linewidth]{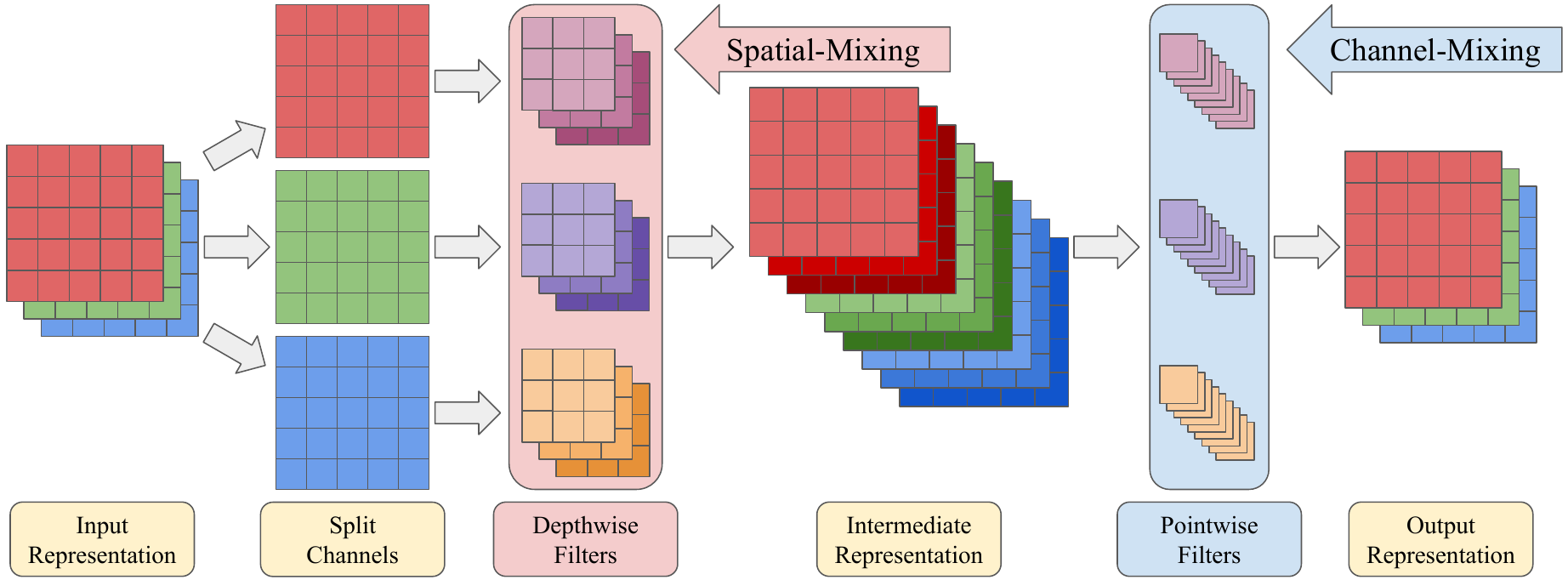}
    \caption{Separable convolutions allow us to disentangle the roles of spatial and channel mixing in deep networks by freezing either the depthwise or pointwise filters and learning the others.}
    \label{fig:static}
    \vspace{-0.3cm}
\end{figure*}

\begin{figure}
    \centering
    \begin{tabular}{cc}
        \includegraphics[width=0.45\linewidth]{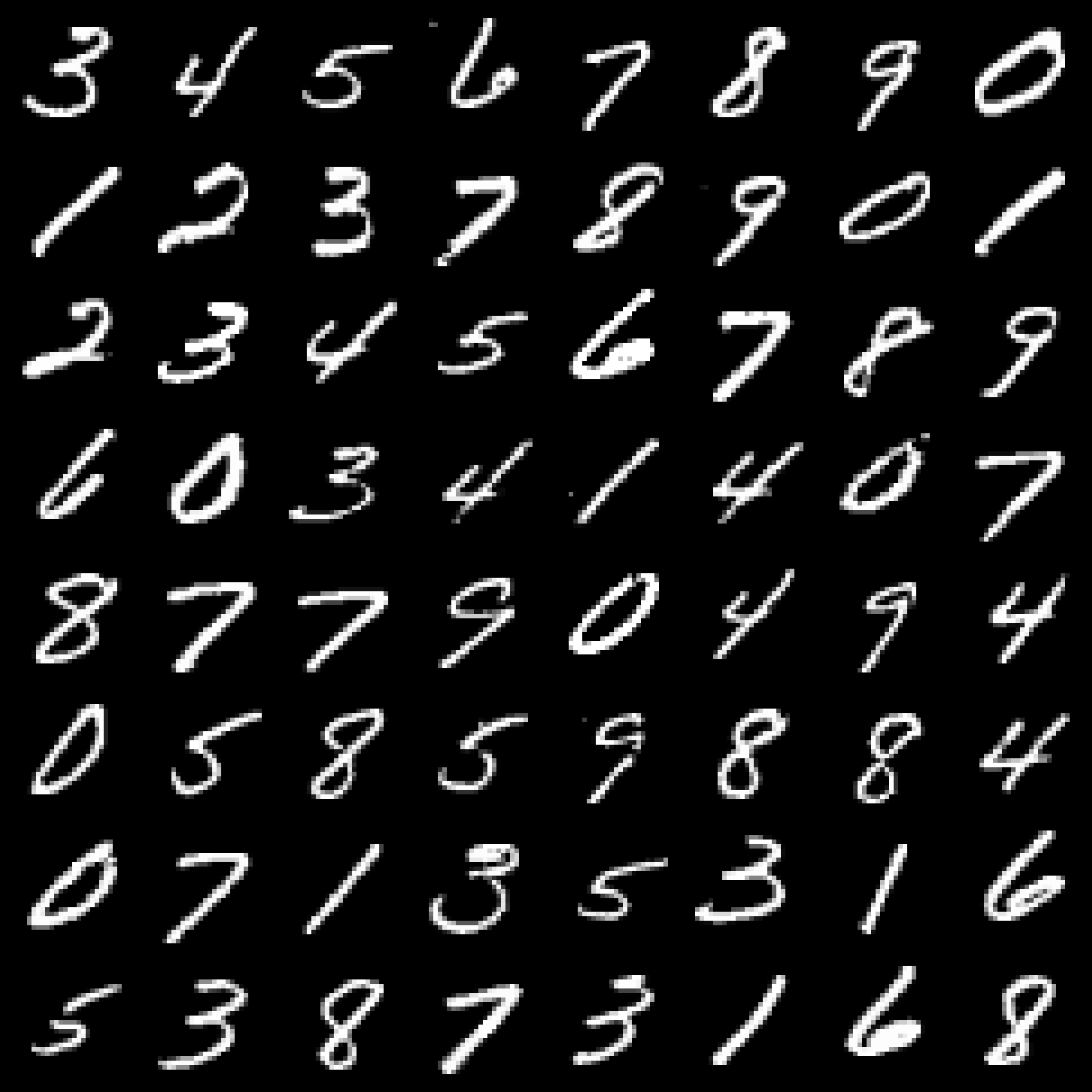} & \includegraphics[width=0.45\linewidth]{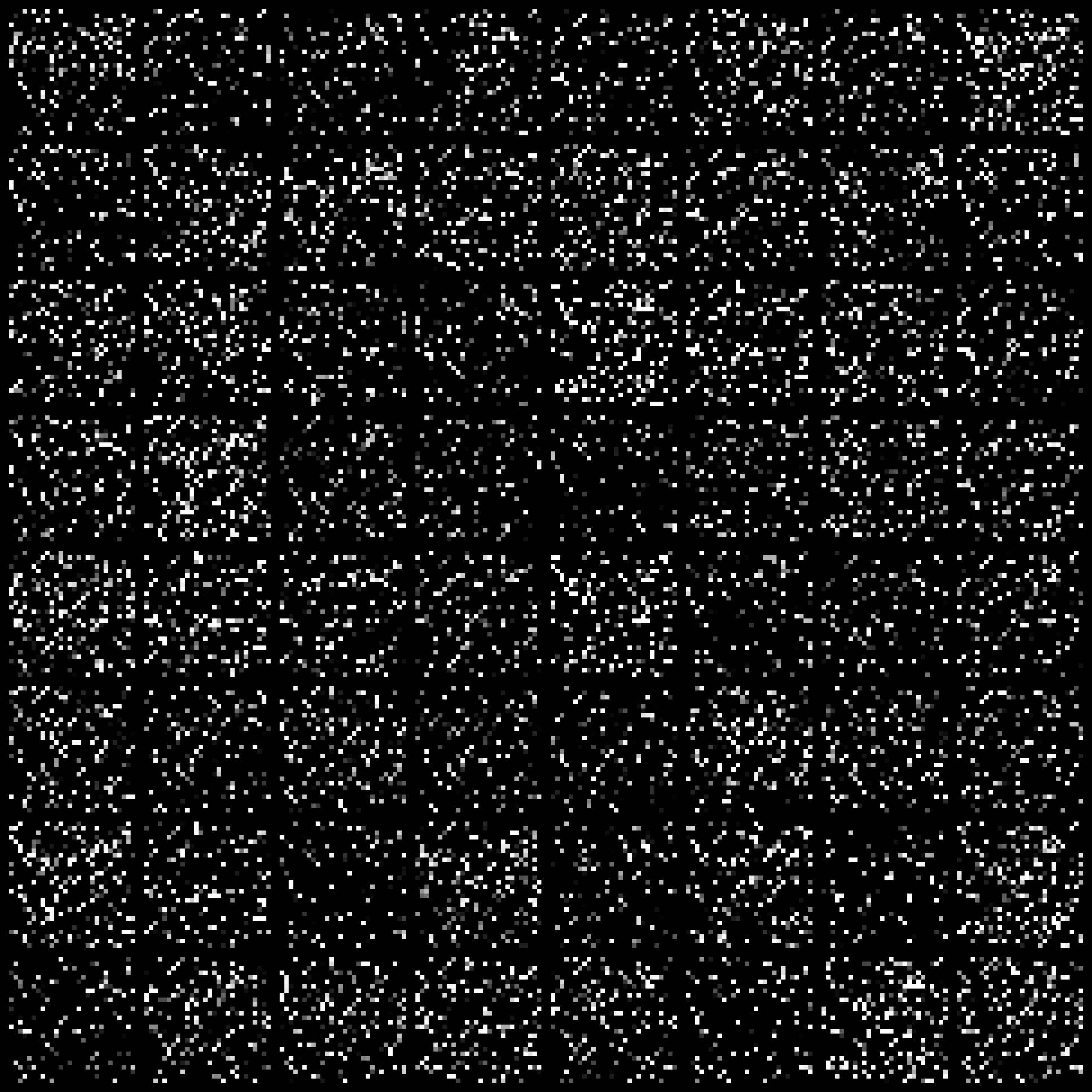} \\
        (A) Ground Truth & (B) Scrambled Pixels \\
        \includegraphics[width=0.45\linewidth]{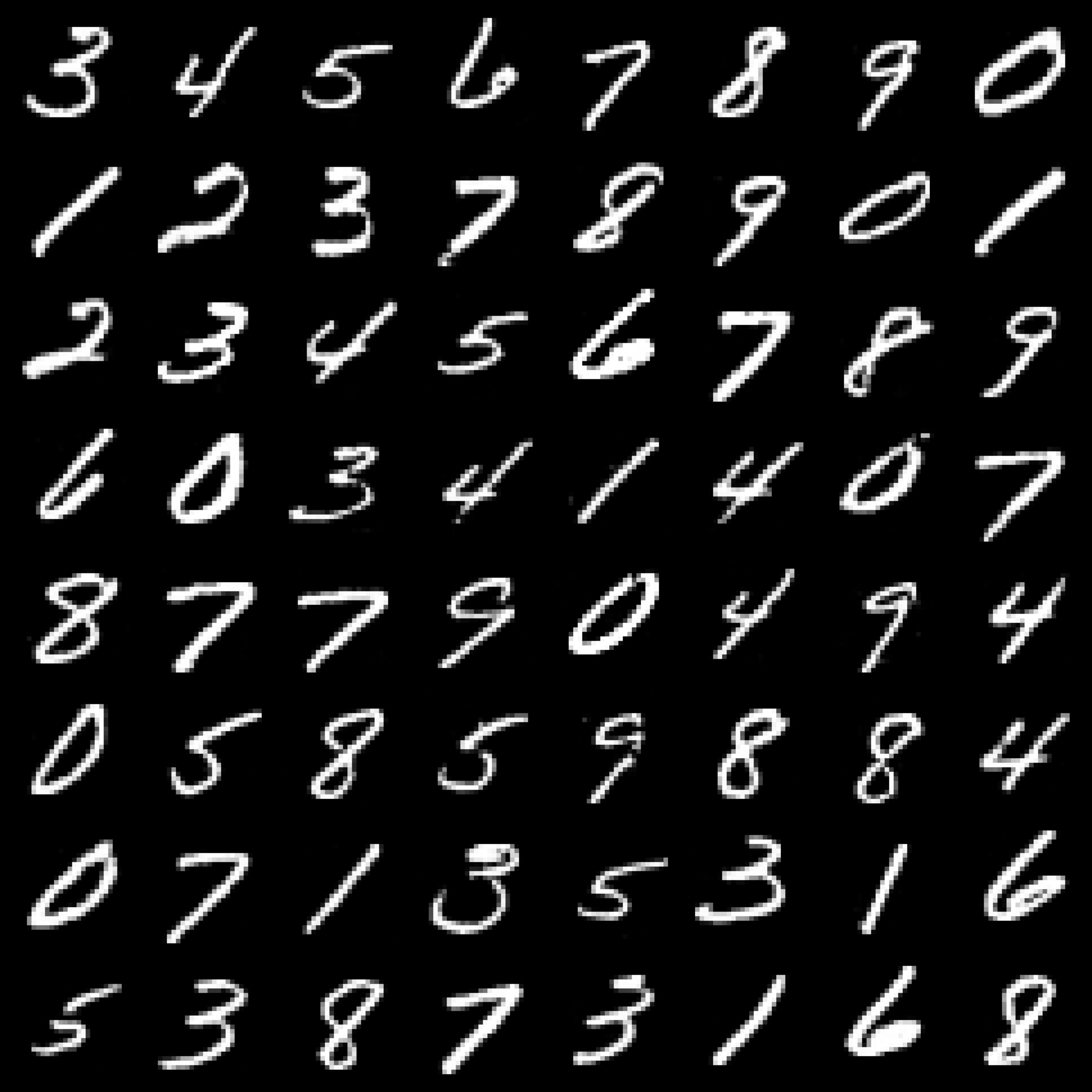} & \includegraphics[width=0.45\linewidth]{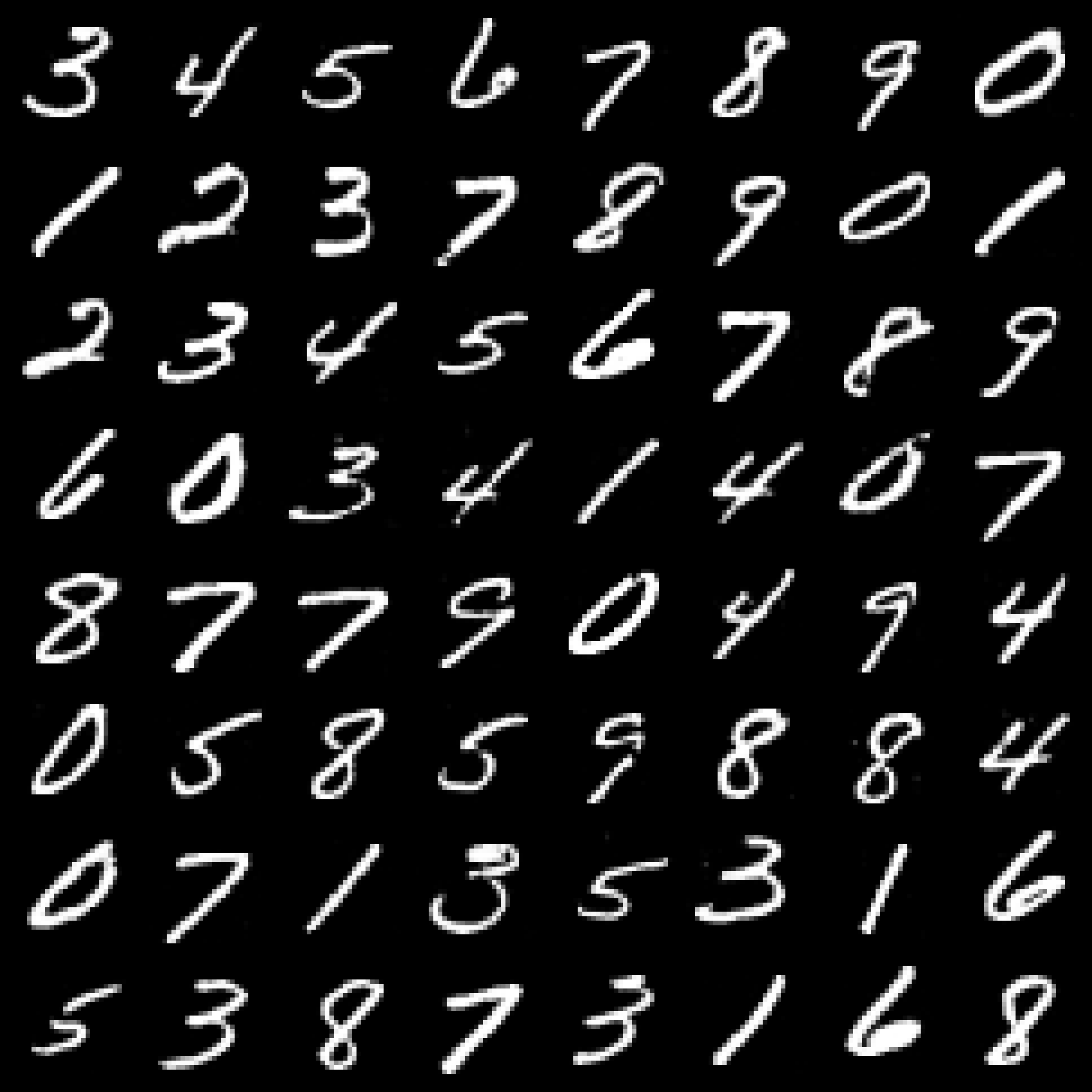} \\
        (C) Reconstruction (Full) & (D) Reconstruction (Chans)
    \end{tabular}
    \caption{Models that only learn channel mixing are capable of learning near perfect reconstruction of MNIST digits from their shuffled pixels.}
    \label{fig:mnist}
\end{figure}\vfill\null
\begin{figure}
    \centering
    \begin{tabular}{cc}
        \includegraphics[width=0.45\linewidth]{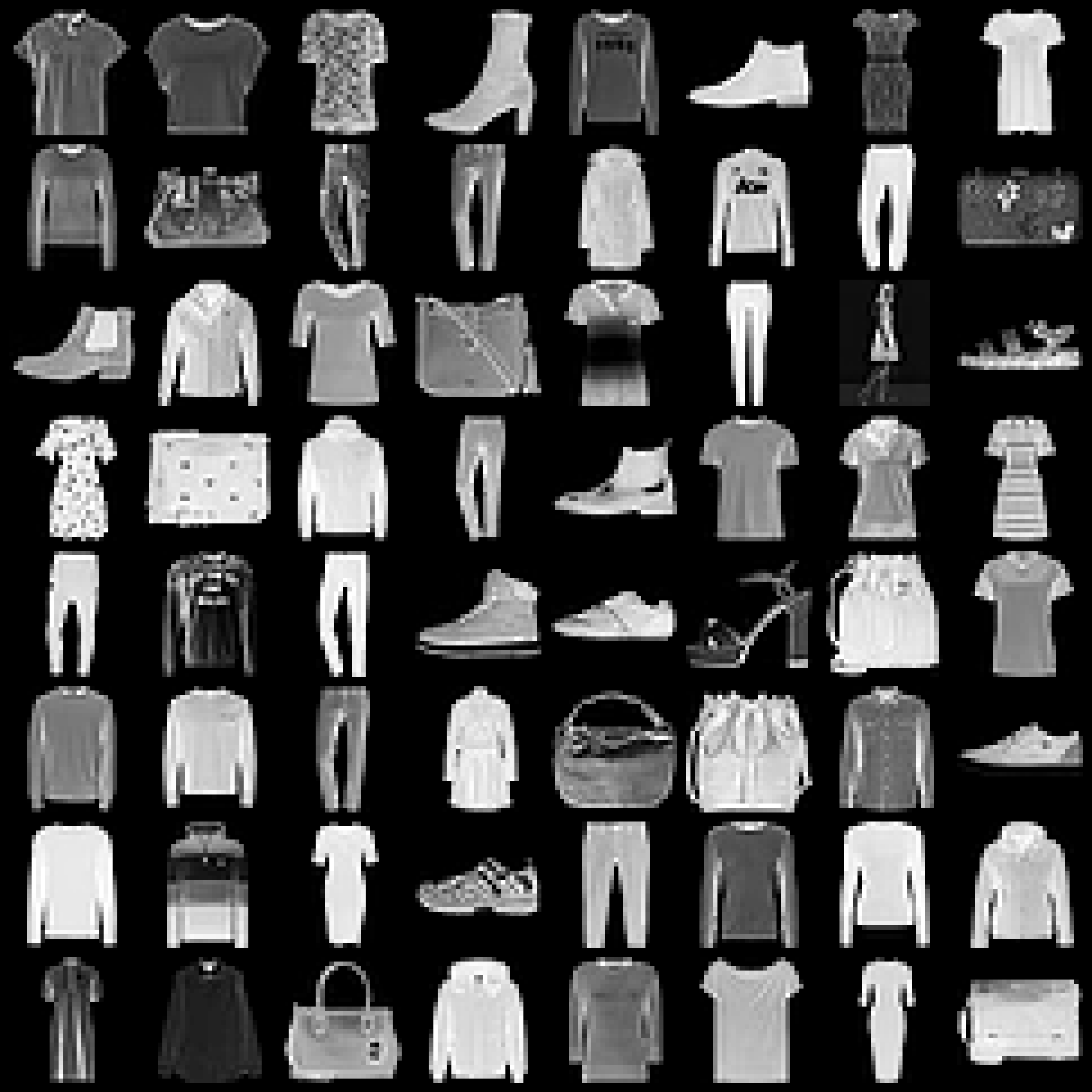} & \includegraphics[width=0.45\linewidth]{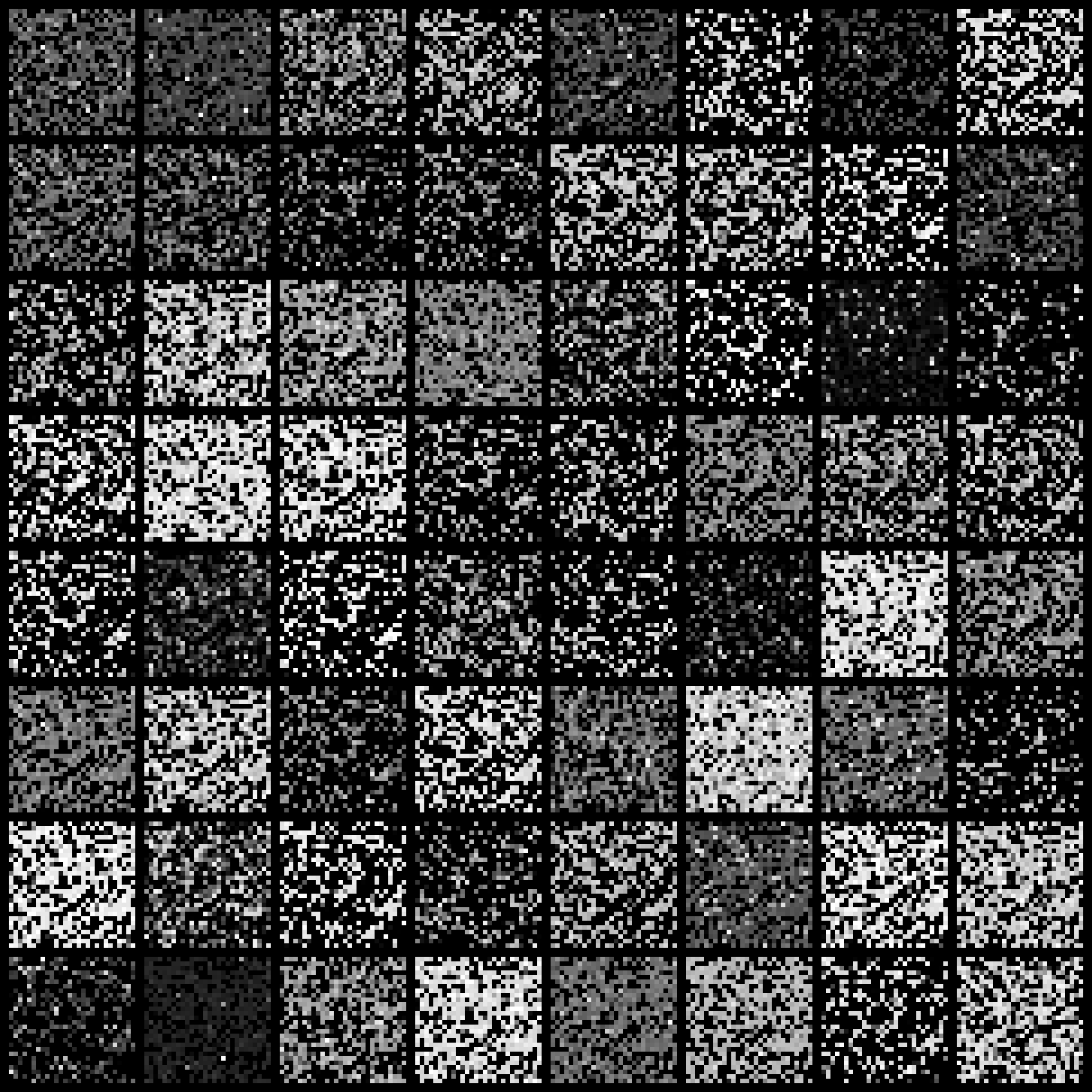} \\
        (A) Ground Truth & (B) Scrambled Pixels \\
        \includegraphics[width=0.45\linewidth]{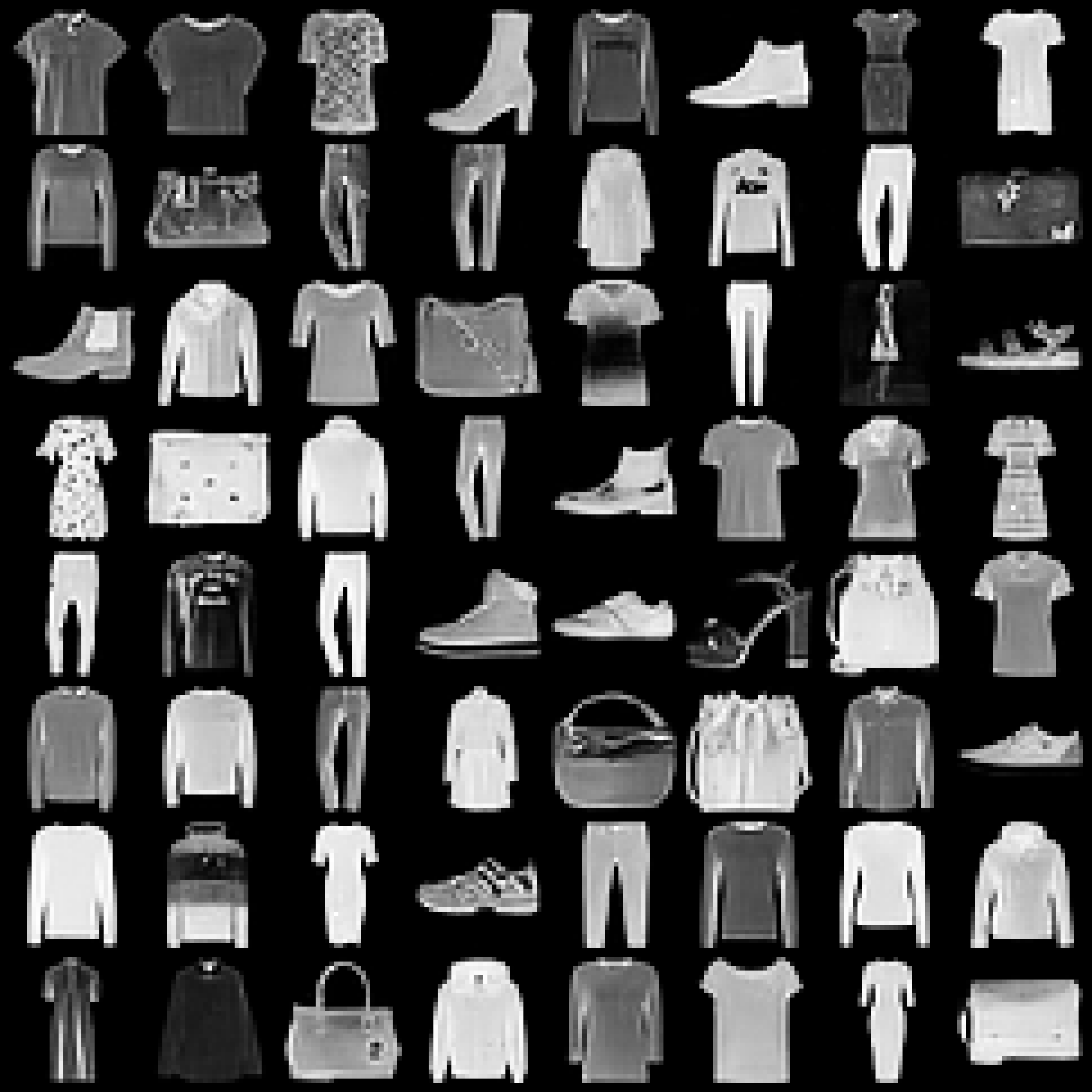} & \includegraphics[width=0.45\linewidth]{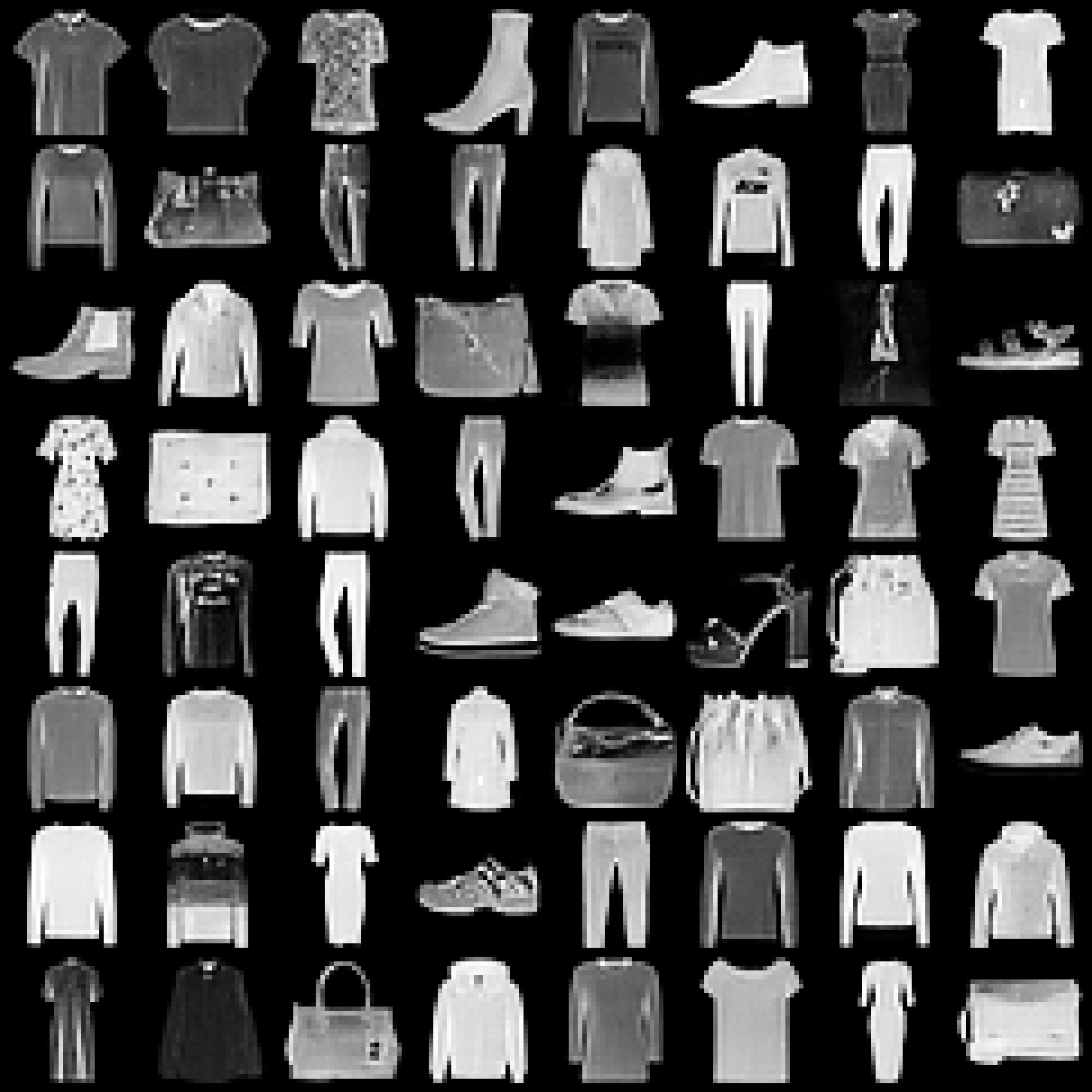} \\
        (C) Reconstruction (Full) & (D) Reconstruction (Chans)
    \end{tabular}
    \caption{Even on MNIST-Fashion, a significantly more complex dataset than MNIST-Digits, the models that learn only channel mixing still achieve near perfect un-shuffling with results almost indistinguishable from those of the fully-learned models.}
    \label{fig:fashion}
\end{figure}

\begin{figure*}[!h]
    \centering
    \begin{tabular}{cccccc}
    & $k=3$ && $k=5$ && $k=7$ \vspace{-0.1cm}\\
        \rotatebox[origin=c]{90}{Test PSNR~~~~~~~~} &\hspace{-.1cm} \includegraphics[trim={0.8cm 0.4cm 0.2cm 0.4cm},clip,align=c,width=0.25\linewidth]{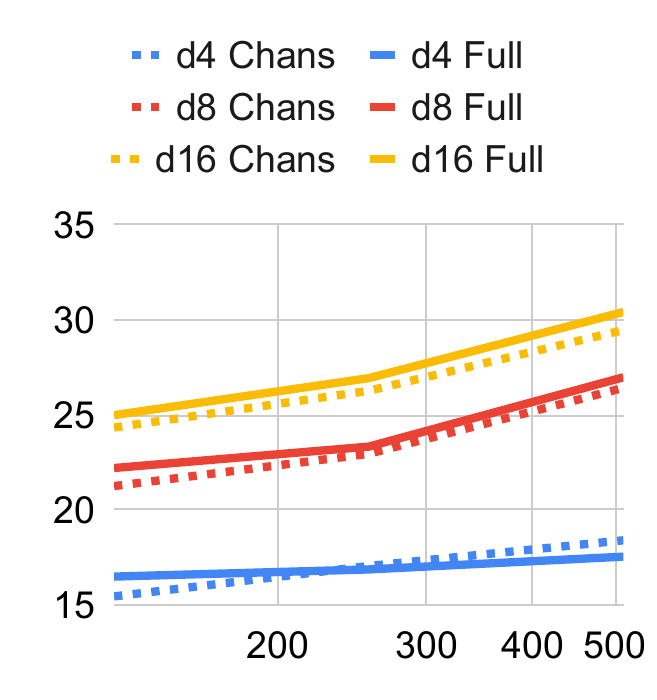} & \rotatebox[origin=c]{90}{Test PSNR~~~~~~~~} & \hspace{-.1cm} \includegraphics[trim={0.8cm 0.4cm 0.2cm 0.4cm},clip,align=c,width=0.25\linewidth]{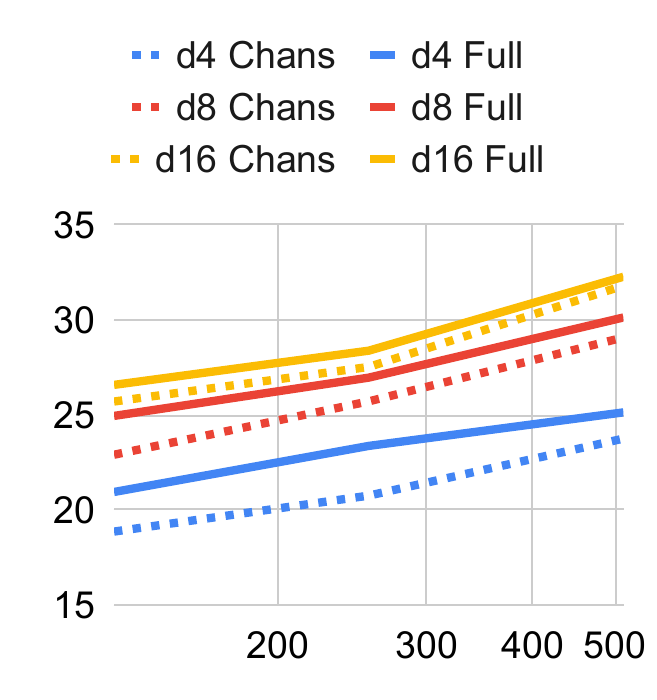} & \rotatebox[origin=c]{90}{Test PSNR~~~~~~~~} &\hspace{-.1cm} \includegraphics[trim={0.8cm 0.4cm 0.2cm 0.4cm},clip,align=c,width=0.25\linewidth]{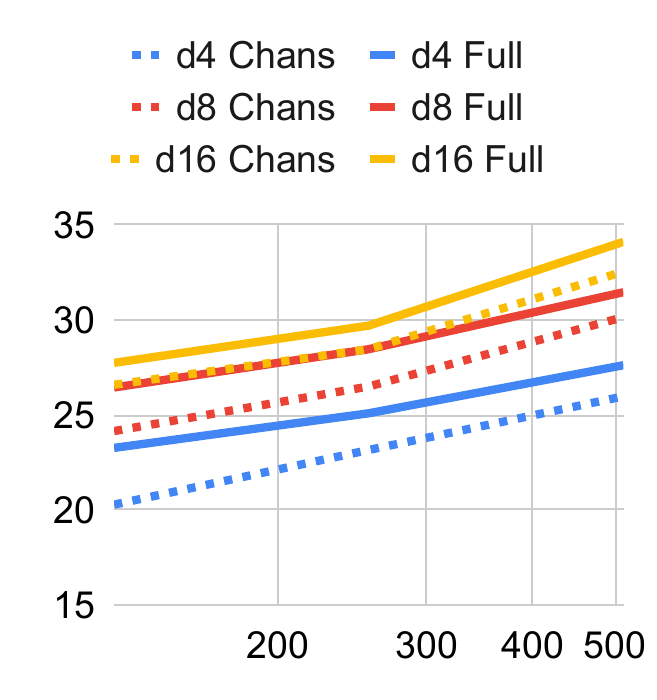} \vspace{-0.cm}\\
        & Width && Width && Width
    \end{tabular}
    \vspace{-0.25cm}
    \caption{\textbf{MNIST}: PSNR for Pixel Un-Shuffling.}
    \vspace{-0.5cm}
    \label{fig:pnsr}
\end{figure*}
\begin{figure*}[!h]
    \centering
    \begin{tabular}{cccccc}
    & $k=3$ && $k=5$ && $k=7$ \vspace{-0.1cm}\\
        \rotatebox[origin=c]{90}{Test PSNR~~~~~~~~} &\hspace{-.1cm} \includegraphics[trim={0.8cm 0.4cm 0.2cm 0.4cm},clip,align=c,width=0.25\linewidth]{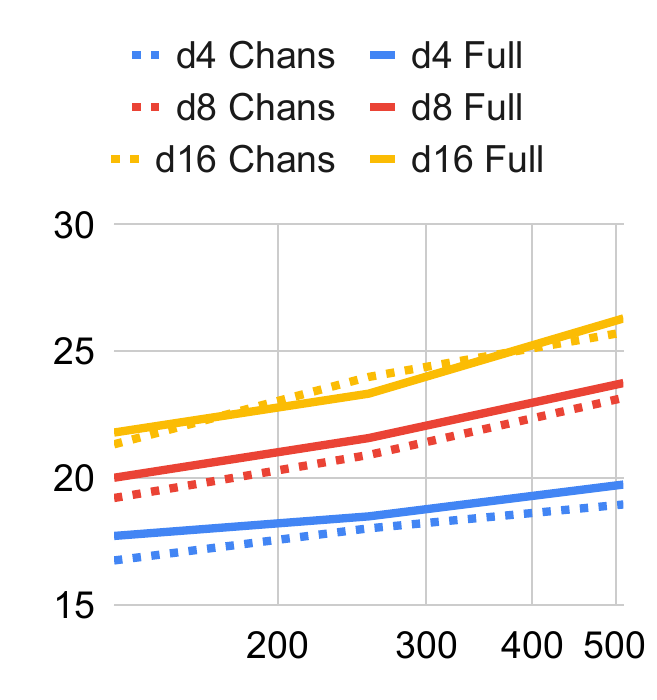} & \rotatebox[origin=c]{90}{Test PSNR~~~~~~~~} & \hspace{-.1cm} \includegraphics[trim={0.8cm 0.4cm 0.2cm 0.4cm},clip,align=c,width=0.25\linewidth]{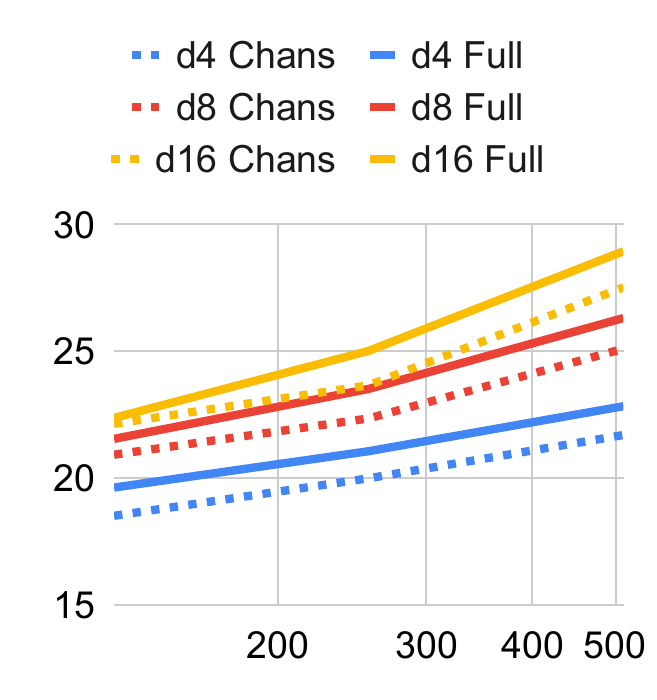} & \rotatebox[origin=c]{90}{Test PSNR~~~~~~~~} &\hspace{-.1cm} \includegraphics[trim={0.8cm 0.4cm 0.2cm 0.4cm},clip,align=c,width=0.25\linewidth]{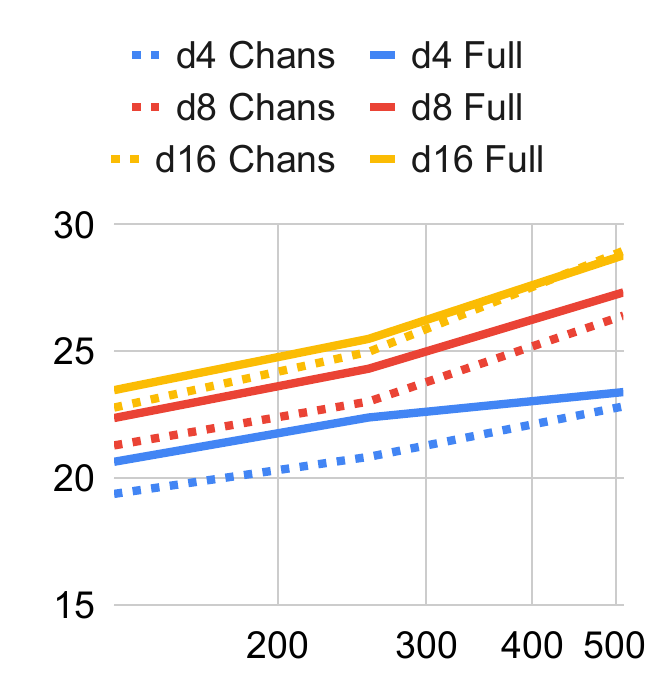} \vspace{-0.cm}\\
        & Width && Width && Width
    \end{tabular}
    \vspace{-0.25cm}
    \caption{\textbf{Fashion-MNIST}: PSNR for Pixel Un-Shuffling.}
    \vspace{-0.5cm}
    \label{fig:pnsr}
\end{figure*}

 \fi
\end{document}